\pdfoutput=1
\documentclass[11pt]{article}

\usepackage{acl}
\usepackage{times}
\usepackage{latexsym}
\usepackage[T1]{fontenc}
\usepackage[utf8]{inputenc}
\usepackage{microtype}
\usepackage{float}
\usepackage{xcolor}
\usepackage{amssymb}
\usepackage{pifont}% http://ctan.org/pkg/pifont
\usepackage{amsmath}
\usepackage{comment}
\usepackage{inconsolata}
\usepackage{graphicx, booktabs, multirow}
\usepackage{makecell}   % nicer line breaks inside cells} % tables
\usepackage{pgfplots}
\pgfplotsset{compat=1.18}
\usepackage{hyperref}
\usepackage{cleveref}
\usepackage[colorinlistoftodos]{todonotes}
\usepackage{listings}
\usepackage{fontawesome5} % Or simply 'fontawesome' if you don't need the latest icons

\newcommand{\githubrepo}[1]{%
    \href{#1}{\small \faGithub \ \texttt{#1}}%
}
\definecolor{darkred}{HTML}{8B0000} % Classic Hex DarkRed
\definecolor{darkgreen}{HTML}{006400} % Hex code for DarkGreen
\newcommand{\cmark}{\textcolor{darkgreen}{\ding{51}}}%
\newcommand{\xmark}{\textcolor{darkred}{\ding{55}}}%
\newtheorem{researchQuestion}{RQ}
\newcommand{\R}{\mathbb{R}}

\newcommand{\bz}{{\boldsymbol{z}}}

\title{ToMMeR -- Efficient Entity Mention Detection from Large Language Models}

\author{
  Victor Morand$^{1}$ \quad
  Nadi Tomeh$^{2}$  \quad
  Josiane Mothe$^{3}$ \quad
  Benjamin Piwowarski$^{1}$ \\
  \\
  $^{1}$Institut des Systèmes Intelligents et de Robotique (ISIR),\\ 
  \vspace{0.2cm}
  Sorbonne Université, CNRS, F-75005 Paris, France \\
  \vspace{0.2cm}
  $^{2}$LIPN, Université Sorbonne Paris Nord, UMR7030 CNRS \\
  $^{3}$INSPE, UT2J, Univ. Toulouse, IRIT, CLLE \\ UMR5505 UMR5263 CNRS, F-31400 Toulouse, France \\
  }

\begin{document}
\maketitle

\begin{abstract}
Identifying which text spans refer to entities --mention detection-- is both foundational for information extraction and a known performance bottleneck.
We introduce ToMMeR, a lightweight model (<300K parameters) probing mention detection capabilities from early LLM layers. Across 13 NER benchmarks, ToMMeR achieves 93\% recall zero-shot, with an estimated 90\% precision under a human-calibrated LLM-judge protocol, showing that ToMMeR rarely produces spurious predictions despite high recall. Cross-model analysis reveals that diverse architectures (14M-15B parameters) converge on similar mention boundaries (DICE >75\%), confirming that mention detection emerges naturally from language modeling.  When extended with span classification heads, ToMMeR achieves competitive NER performance (80-87\% F1 on standard benchmarks). Our work provides evidence that structured entity representations exist in early transformer layers and can be efficiently recovered with minimal parameters.
% TODO 

\end{abstract}

\begin{center}
    \githubrepo{https://github.com/VictorMorand/llm2ner}
\end{center}

\section{Introduction}\label{sec:introduction}

Information extraction (IE) pipelines start with a fundamental task: \emph{mention detection}—identifying text spans that refer to entities or concepts worth tracking. These mentions range from specific referential entities (\emph{Marie Curie}, \emph{Tesla Inc.}) to abstract concepts (\emph{philosopher}, \emph{oxidation process}), typically realized as noun phrases. Despite its importance, mention detection remains a recognized bottleneck in NER systems \cite{popovicEmbeddedNamedEntity2024}, yet it is almost always conflated with entity typing in a single joint task. This conflation obscures a key question: \emph{Where and how do models learn to detect span boundaries?}

In contrast with entity labels, mention boundaries are fundamentally schema-invariant, decoupling detection from typing could then be key to building IE systems that transfer better. While NER models  typically employ hundreds of millions of parameters trained on task-specific annotations \cite{zaratianaGLiNERGeneralistModel2023}, evidence from mechanistic interpretability suggests that Large Language Models (LLMs) may already encode entity spans during pretraining \cite{FengSteinhardt2024, gevaDissectingRecallFactual2023, morand2025representationsentitiesautoregressivelarge}. If mention detection emerges from language modeling objectives, \emph{it should be recoverable from LLM representations with minimal additional parameters}.

\begin{figure}[t]
    \centering
    \includegraphics[width=\linewidth]{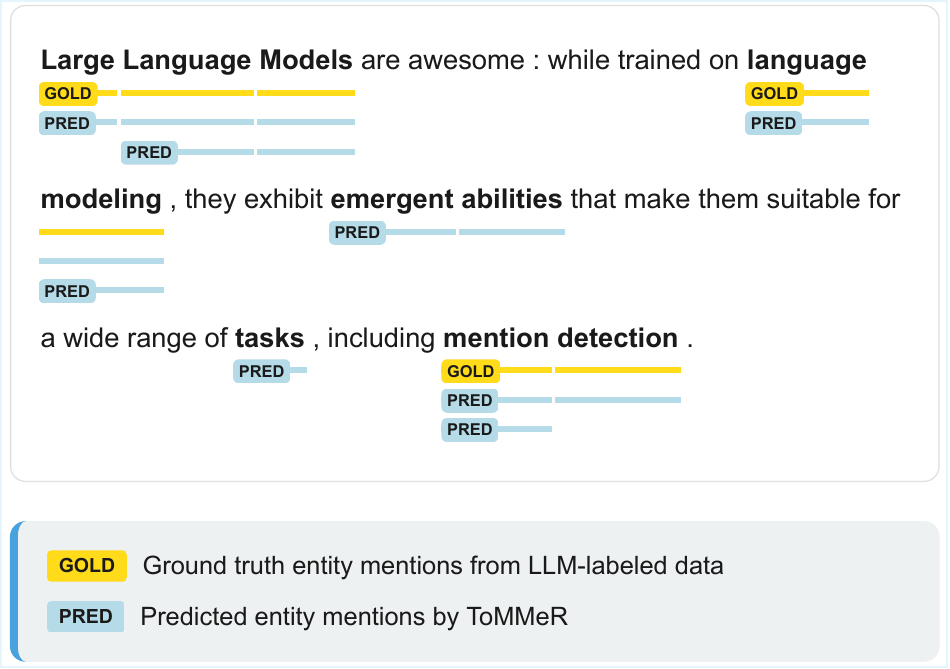}
    \caption{ToMMeR, a lightweight model probing  emergent mention detection capabilities from early layers representations of any LLM backbone. Trained to generalize LLM annotated data, ToMMeR achieves high Zero-Shot recall across a wide set of NER benchmarks.}
    \label{fig:AbstractFigure}
    \vspace{-0.5cm}
\end{figure}

We propose ToMMeR (Token Matching for Mention Recognition), a lightweight architecture (under 300K parameters, trainable in hours) that scores spans from early layers of frozen LLM backbones, using only one (partial) forward pass—no prompting, no schema specification, no text generation (\textit{$42 \times$ faster than prompting methods}, cf App \labelcref{app:Efficiency}). We train ToMMeR using only span \emph{boundaries} from Pile-NER—GPT-3.5 annotations on samples from \emph{The Pile}~\citep{zhou2023pilener,gao2020pile}. Because typed NER benchmarks under-label many legitimate mentions (nested, generic, etc.), ``precision'' against gold often penalizes coverage, we thus complement initial benchmark annotation by estimating ToMMeR's precision with human-calibrated LLM judges. Across 13 NER benchmarks, ToMMeR achieves 93\% recall with precision estimated over 90\%, while also showing strong multi-lingual transfer on Latin scripts (\Cref{tab:ZeroShotResults}). Cross-model analysis reveals that diverse LLM architectures (14M to 15B, both auto-regressive and encoder-only) converge on similar mention boundaries (DICE scores $>$0.75), suggesting mention detection is a shared, emergent capability rather than a dataset artifact.

While various systems perform untyped mention detection, most rely on supervision tied to specific datasets or annotation schemes. Coreference models rank spans but inherit conventions from their training data~\citep{lee2017end}. Weakly supervised approaches prioritize high-recall proposals when gold annotations are incomplete~\citep{werlen2020partially}, and span-based event systems detect untyped triggers before clustering~\citep{lu2021span}. In all cases, span scorers remain shaped by benchmark-specific schemas and do not transfer cleanly across domains or annotation guidelines.
Closer to our work, EMBER~\citep{PopovicFaerber2024} trains NER models over LLM attention scores and hidden states, but requires labeled data for each schema, thus remaining tailored to specific datasets. Generalist extractors such as GLiNER broaden coverage, supporting zero/low-shot transfer, yet require an input schema at inference time, reintroducing task specification and alignment costs~\citep{zaratiana-etal-2024-gliner}.

Our contributions are threefold: \textbf{(i)} a simple, efficient probing architecture that recovers mention spans from a single forward pass of early layers; \textbf{(ii)} empirical evidence that mention boundaries are robustly encoded across layers, models, scales, and architectural families, with consistent cross-model predictions despite no shared supervision; and \textbf{(iii)} we release ToMMeR models and demonstrate a straightforward extension to full NER via span classification, achieving competitive performance (80-87\% F1) on standard benchmarks and enabling modular, schema-agnostic extraction pipelines.

\begin{figure}[t]
    \centering
    \includegraphics[width=0.9\linewidth]{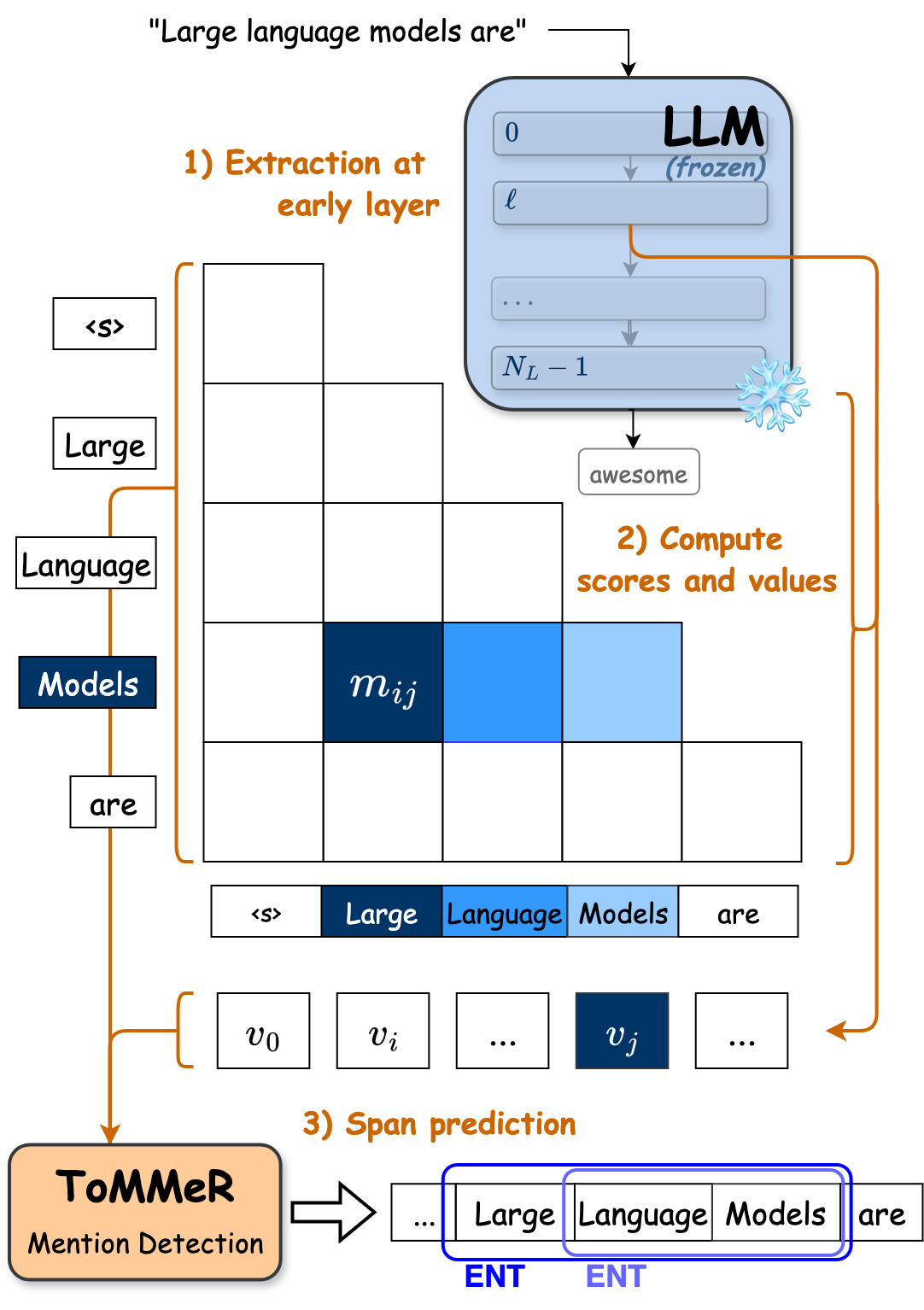}
    \caption{The ToMMeR architecture. We extract the mention detection capabilities of any \textit{frozen} LLM backbone with less than $300$K additional parameters, computationally equivalent to an additional attention head. We leverage \textit{Matching scores} $m_{ij}$  between tokens $t_i$ and $t_j$ and individual values $v_{i}$, all probed from LLM representations at layer $\ell$.}
    \label{fig:Architecture}
    \vspace{-0.4cm}
\end{figure}

\section{ToMMeR} \label{sec:Tommer}
\paragraph{Entity Mentions in Transformers.}
In transformer-based language models \cite{vaswaniAttentionAllYou2017}, a text is tokenized into a sequence of tokens $(t_1, \ldots , t_n) \in \mathcal V^n$, with $\mathcal V$ the vocabulary used by the tokenizer. These tokens are embedded into a sequence of initial \textit{representations} that are sequentially processed through the transformer layers. Each layer $\ell \in \{0,\ldots,N_L-1\}$ generates a new series of {representations} $(\bz_1^\ell, \ldots, \bz_n^\ell) \in \mathbb{R}^d$ from the representations of the preceding layer. 

For each sentence  $(t_1, \ldots , t_n) \in \mathcal V^n$, we consider the set of $N_E$ \textit{entity mentions} $E = \left\{(s_k , e_k) \in [1,n]^{2 \times N_E}\right\}$ with $s_k$ and $e_k$ respectively the start and end \textit{token} indexes of entity mention $k$, constrained in this work to contiguous spans of length $\leq 25$ tokens, covering the majority of mentions\footnote{More details on the dataset in \Cref{app:PileNERStats}}. This limit reduces the quadratic complexity of span enumeration while preserving most linguistically valid mentions. The task is then framed as binary classification: for each span, determine whether it constitutes a valid mention.

\paragraph{Architecture.}
\label{subsec:TommerArchitecture}
We design ToMMeR as a probing classifier—a small neural network trained on frozen representations to recover latent capabilities.  Unlike fine-tuning, probing preserves the backbone model and requires minimal parameters.
More specifically, we ground our approach on the binding ID framework from mechanistic interpretability \cite{fengHowLanguageModels2024}, which posits that transformers dynamically bind related tokens through learned signals, enabling later retrieval via attention. Extending this idea, we hypothesize that LLMs implicitly group entity-mention tokens using analogous binding mechanisms---effectively encoding mention boundaries within their hidden states. ToMMeR leverages these latent binding signals to extract mention spans directly from representations of a frozen LLM, \textit{requiring no modifications to the backbone model}. For each pair $(\bz_i^\ell ,\  \bz_j^\ell)_{1 \leq i \leq j \leq n}$ of token representations at layer $\ell$, the \textit{Matching score} $m_{ij} \in \mathbb R$ quantifies this association.

\paragraph{Matching Score.}\label{sec:ToM}
To detect entity bindings, we adapt the transformer’s attention mechanism, which measures token similarity via dot products between query and key vectors. While standard attention computes a probability distribution over tokens, our goal is to capture binary token-to-token matching. Thus, we replace softmax with $\ell_2$ normalization which proved to be stable across backbones, yielding cosine similarity as our matching metric. Formally, we compute pairwise scores using learned projections on a rank $r$ query-key subspace $(W_Q, W_K) \in \mathbb{R}^{r \times d}$. 

\vspace{-8pt}
\begin{equation}
\label{eq:TM_qk}
    m_{ij} = \cos ( W_Q \bz_i^\ell \ , \  \underbrace{W_K \bz_j^\ell}_{\in \mathbb R^r} ) \in \mathbb [0,1]
\end{equation}
\vspace{-7pt}

We also explore other normalization functions and formulations for the matching score, probing different transformer components in Appendix \labelcref{app:OtherArchis}.

\paragraph{Token Values and Span Probability.}
To complement pairwise matching scores, which capture inter-token bindings but lack boundary and anchor information, we incorporate token-level information  with a learned linear layer (or probe) $v_{i} = W_V \bz_i^\ell \in \mathbb R$. These values leverage the observation that LLMs concentrate entity information in its final token representation~\cite{mengLocatingEditingFactual2022, gevaDissectingRecallFactual2023}, providing critical cues for autoregressive models. The final span probability $\hat{p}_{ij} \approx p \left((i,j) \in E \right)$ is predicted with a logistic model, with parameters $\theta \in \R^5$, and inputs the matching scores and individual values around the mention's last token. The model is given by equ.~\ref{eq:Combine_scores}.

\vspace{-8pt}
\begin{equation}
\label{eq:Combine_scores}
    \hat p_{ij} = \theta \cdot
    \begin{pmatrix}
        m_{ij} \\
        \max\{m_{kj}\}_{i<k \leq j} \\
        \min\{m_{kj}\}_{i<k \leq j} \\
        v_j \\
        v_{j+1}
    \end{pmatrix}
\end{equation}
\vspace{-7pt}

\noindent where  max/min pooling  over intermediate matching scores $m_{ik}$ captures the strength of internal token bindings within the span, and $v_j, v_{j+1}$ provide information about the span's last token and its immediate context (See architecture  \Cref{fig:Architecture}).

\subsection{Training on Span Detection}

\paragraph{Data.}\label{par:Dataset}

We use Pile-NER \cite{zhouUniversalNERTargetedDistillation2023}, a dataset of 45,889 samples from The Pile \cite{gao2020pile800gbdatasetdiverse}, annotated with fine-grained entity types using GPT-3.5. While originally designed for zero-shot NER systems like UniversalNER \cite{zhouUniversalNERTargetedDistillation2023} and GLiNER \cite{zaratianaGLiNERGeneralistModel2023}, its broad semantic coverage and diverse mention types make it ideal for studying mention detection. Additional dataset statistics are provided in Appendix~\ref{app:AdditionnalFigs}.

\paragraph{Loss.}\label{par:Loss}
ToMMeR's parameters are optimized end-to-end on binary span classification (valid/invalid). Mention detection however faces severe class imbalance, as non-entity spans (negative examples) vastly outnumber entity mention spans (positive examples), even in Pile-NER. To address this, we employ Balanced Binary Cross-Entropy (BBCE), which reweights the standard BCE loss using a dynamic factor $\alpha$. This ensures equal contribution from both classes regardless of their imbalance:
For each batch, the loss is computed as:

\vspace{-0.3cm}
\begin{equation}
\begin{aligned}
    \texttt{BBCE}(\hat p, y) = \frac{-1}{\#_\text{Tot}}\sum_{i<j}  \alpha \ y_{ij} \ \log(&\hat p_{ij}) \\
                        + \ (1 - y_{ij})\ \log(1 - &\hat p_{ij})
\end{aligned}
\end{equation}
where $\hat p_{ij}$ is the predicted probability, $y_{ij}$ is the gold binary label,  $\alpha $ is the balanced class weight, computed for each batch as $\alpha = \frac{\#\text{Neg}}{\#\text{Pos}}$, and $\#_\text{Tot}$ is the total number of spans in the batch.

\paragraph{Distillation.}\label{par:Distillation}
Although it has been generated with an LLM, and already contains an important number of fine-grained entity types, \hyperref[par:Dataset]{Pile-NER} also suffers from incompleteness. Nested mentions are for instance not labeled. Moreover, even when mentions are extracted by the LLM, the labels may not fully reflect the internal notion of mention detection.
To mitigate these limitations, we adopt a two-stage training strategy: after a first fit on the available annotations, we use the learned model to augment the training dataset with new mentions that were not annotated in the data, thus reducing the number of false negatives.
See Appendix \labelcref{app:Reproducibility} for detailed hyper-parameters.

\section{Related Work} \label{sec:related}

\paragraph{Mechanistic and Algorithmic Structure.}
Mechanistic studies suggest specific circuits for entity-related behavior: \emph{Binding ID} capture abstract entity via activation directions \citep{FengSteinhardt2024}; induction heads implement copy/coreference-like mechanisms \citep{Olsson2022InductionHeads}; attention heads bracket NPs or track antecedents \citep{Clark2019WhatDoesBERTLookAt}; and modular subgraphs compose across subtasks \citep{mondorf-etal-2025-circuit}. Complementarily, probing/theoretical work recovers tree or chart-like structure from hidden states (e.g., Inside-Outside/CKY signals) \citep{Zhao2023InsideOutside,Tenney2019EdgeProbing}, and \citet{AllenZhuLi2023Physics} show causal autoregressive models can learn formal grammars, with hidden states linearly encoding boundaries and attention flows mimicking dynamic programming. Recent work also studies entity identification directly in language-model representations, showing that entity information is highly linearly separable and concentrated in low-dimensional subspaces of early layers \citep{sakata-etal-2025-entity}. These findings support the view that the allocation of probability over spans is a natural byproduct of next-token prediction; we connect this to the detection of untyped mentions extracted from early layers.

\paragraph{Embedded/Probing Detectors and Span Boundary Models.}
Low-latency extraction can be achieved with probes on frozen LLMs. \citet{PopovicFaerber2024} predict token and span-boundary signals from hidden states and attention during generation. While EMBER targets schema-specific NER with supervised training, we extract a schema-agnostic notion of mention that generalizes zero-shot. Classic probing shows entity/span information concentrates in intermediate layers, and structural probes reveal linear syntax/span structure \citep{Tenney2019EdgeProbing,HewittManning2019}; probes have also tested entity state tracking \citep{kim2023entitytracking}. Orthogonally, pointer/boundary decoders focus on detecting mention as the bottleneck for Information Extraction/Entity Linking \citep{Li2019PointerNER,shang-etal-2018-learning,bian2023dmner}.

\paragraph{Open-Schema IE and Generalist NER.}
A broad line of work targets ontology-agnostic IE via generalist or instruction-driven interfaces. GLiNER and GLiNER2 support schema-driven extraction and zero/low-shot transfer \citep{zaratiana-etal-2024-gliner,zaratiana2025gliner2}; UniversalNER distills LLM capabilities into smaller models for open NER \citep{Zhou2023UniversalNER}. Unified text-to-structure frameworks and instruction-tuned systems (UIE, USM, InstructUIE, RAIT, YAYI-UIE, PIVOINE, RUIE, InstructIE, TRUE-UIE) expand this paradigm with retrieval and prompting strategies \citep{lu-etal-2022-unified,lou2023usm,wang2023instructuie,xie2025rait,xiao2024yayi,lu2023pivoine,liao2025ruie,jiao2023instructie,wang2024trueuie}. Recent LLM-based NER work also improves in-context extraction without parameter updates through label-guided demonstration retrieval and targeted error reflection \citep{bai-etal-2025-llms}. Retrieval-based mention retrieval further enables zero-shot typing \citep{shachar2025nerretriever}. We differ by \emph{probing} early layers of frozen LLMs to recover a model-internal notion of entity mention with minimal additional parameters, rather than training a new generalist encoder or relying on in-context demonstrations at inference time.

\paragraph{Distillation, Pseudo-Labels, and Evaluation.}
Distillation from LLM to small models aids broad-coverage information extraction \citep{zhouUniversalNERTargetedDistillation2023}; self-training and confidence-based pseudo-labeling mitigate annotation gaps \citep{Sohn2020FixMatch}, alongside prototype/contrastive approaches and pseudo-label refinement \citep{zhou-etal-2023-improving,zhang2023taskrelation}. Relatedly, weakly supervised few-shot domain adaptation methods leverage small labeled support sets together with unlabeled target-domain data, for instance through joint constrained k-means and discriminative subspace selection for NER \citep{hammal-etal-2025-shot}. Our approach complements these methods by revealing that LLMs already encode a rich, generalizable notion of entity mentions—effectively distilling and amplifying this latent knowledge with minimal parameter overhead.

\section{Mention detection Task Evaluation}

\begin{table*}
\centering
\vspace{-.6cm}
\resizebox{0.85\textwidth}{!}{%
\begin{tabular}{@{}crccccccccc@{}}
    &                      & \multicolumn{3}{c|}{\textbf{Threshold decoding}} & \multicolumn{3}{c}{\textbf{\hyperref[sec:decoding]{Greedy} (flat) decoding}} &                      &                     &                 \\                      &                      &                 \\
    & \textbf{Dataset}      & R           & \textcolor{gray}{P}         & \textcolor{gray}{F1}        & R & \textcolor{gray}{P}        & \textcolor{gray}{F1}         & \textbf{\#samples} & \textbf{\#entities} & \textbf{Nested} \\ \midrule
    \multirow{15}{*}{\rotatebox{90}{\textbf{Gold Benchmarks (en)}} }     
    & MultiNERD            & 98.6  & \textcolor{gray}{21.7}   & \textcolor{gray}{35.5}   & 94.0   & \textcolor{gray}{30.0}   & \textcolor{gray}{45.5}  & 154 144 & 23,8005   & \xmark  \\
    & CoNLL 2003           & 94.8  & \textcolor{gray}{33.6}   & \textcolor{gray}{49.6}   & 86.4   & \textcolor{gray}{44.7}   & \textcolor{gray}{58.9}  & 16 493  & 3,4761    & \xmark  \\
    & CrossNER politics    & 97.0  & \textcolor{gray}{32.4}   & \textcolor{gray}{48.6}   & 84.2   & \textcolor{gray}{54.5}   & \textcolor{gray}{66.5}  & 1 389   & 8,838     & \xmark  \\
    & CrossNER AI          & 97.0  & \textcolor{gray}{35.0}   & \textcolor{gray}{51.5}   & 87.2   & \textcolor{gray}{51.2}   & \textcolor{gray}{64.5}  & 879     & 3,776     & \xmark  \\
    & CrossNER literature  & 94.4  & \textcolor{gray}{40.3}   & \textcolor{gray}{56.5}   & 85.9   & \textcolor{gray}{56.6}   & \textcolor{gray}{68.3}  & 916     & 4,749     & \xmark  \\
    & CrossNER science     & 95.7  & \textcolor{gray}{38.2}   & \textcolor{gray}{54.6}   & 85.9   & \textcolor{gray}{55.1}   & \textcolor{gray}{67.1}  & 1 193   & 6,318     & \xmark  \\
    & CrossNER music       & 95.5  & \textcolor{gray}{44.1}   & \textcolor{gray}{60.3}   & 86.7   & \textcolor{gray}{61.7}   & \textcolor{gray}{72.1}  & 945     & 6,420     & \xmark  \\
    & ncbi                 & 91.9  & \textcolor{gray}{12.7}   & \textcolor{gray}{22.2}   & 66.0   & \textcolor{gray}{17.1}   & \textcolor{gray}{27.2}  & 3 952   & 6,808     & \xmark  \\
    & FabNER               & 73.6  & \textcolor{gray}{30.1}   & \textcolor{gray}{42.8}   & 49.0   & \textcolor{gray}{39.7}   & \textcolor{gray}{43.9}  & 13 681  & 64,761    & \xmark  \\
    & WikiNeural           & 97.8  & \textcolor{gray}{20.7}   & \textcolor{gray}{34.1}   & 90.8   & \textcolor{gray}{28.2}   & \textcolor{gray}{43.1}  & 92 672  & 149,005   & \xmark  \\
    & Ontonotes            & 73.0  & \textcolor{gray}{25.5}   & \textcolor{gray}{37.8}   & 59.0   & \textcolor{gray}{31.4}   & \textcolor{gray}{40.9}  & 42 193  & 103,956   & \xmark  \\
    & ACE 2005             & 42.0  & \textcolor{gray}{28.7}   & \textcolor{gray}{34.1}   & 32.5   & \textcolor{gray}{31.9}   & \textcolor{gray}{32.2}  & 8 230   & 30,778    & \cmark  \\
    & GENIA NER            & 95.7  & \textcolor{gray}{24.8}   & \textcolor{gray}{39.4}   & 72.0   & \textcolor{gray}{34.6}   & \textcolor{gray}{46.6}  & 16 563  & 55,968    & \cmark  \\\midrule
    & Aggregated           & 92.6  & \textcolor{gray}{23.2}   & \textcolor{gray}{37.1}   & 84.0   & \textcolor{gray}{31.2}   & \textcolor{gray}{45.5}  & 353 250 & 714,143   &         \\
    & Average              & 88.2  & \textcolor{gray}{29.8}   & \textcolor{gray}{42.6}   & 75.3   & \textcolor{gray}{41.3}   & \textcolor{gray}{52.1}  & 353 250 & 714,143   &         \\\midrule
    \multirow{5}{*}{\rotatebox{90}{\textbf{WikiANN}}}
    & WikiANN - en         & 84.2  & \textcolor{gray}{33.4}   & \textcolor{gray}{47.8}   & 75.5   & \textcolor{gray}{46.2}   & \textcolor{gray}{57.3}  & 40 000  & 56 035     & \xmark \\
    & WikiANN - es         & 85.3  & \textcolor{gray}{34.9}   & \textcolor{gray}{49.5}   & 74.1   & \textcolor{gray}{53.0}   & \textcolor{gray}{61.8}  & 40 000  & 49 280     & \xmark \\
    & WikiANN - fr         & 86.8  & \textcolor{gray}{38.5}   & \textcolor{gray}{53.4}   & 75.8   & \textcolor{gray}{56.1}   & \textcolor{gray}{64.5}  & 40 000  & 52 972     & \xmark \\
    & WikiANN - de         & 91.3  & \textcolor{gray}{33.8}   & \textcolor{gray}{49.3}   & 83.6   & \textcolor{gray}{48.6}   & \textcolor{gray}{61.5}  & 40 000  & 55 329     & \xmark \\
    & WikiANN - zh         & 40.7  & \textcolor{gray}{6.3}    & \textcolor{gray}{10.9}   & 24.0   & \textcolor{gray}{8.8}    & \textcolor{gray}{12.9}  & 40 000  & 50 033     & \xmark \\\midrule  
    \multirow{2}{*}{\rotatebox{90}{\textbf{LLM}}}
    & MultiNERD-gpt-4.1-mini & 71.1  & 92.8 & 80.5   & 50.7   & 95.9   & 66.3 & 1064    & 8,915      & \cmark \\
    & GENIA-gpt-4.1-mini     & 68.4  & 92.3 & 78.6   & 37.5   & 94.9   & 53.7 & 512     & 8,704      & \cmark \\
\bottomrule
\end{tabular}%
}
\caption{Zero-shot mention detection performance of ToMMeR (plugged at layer $6$ of \textsc{Llama3.2-1B}, $274$K parameters only), on various NER benchmarks. \textcolor{gray}{Precision (P)}, Recall (R) and \textcolor{gray}{F1-scores} for threshold and greedy (flat) decoding. \textcolor{gray}{Precision and F1 in gray} stress that low precision is expected when evaluating a schema-agnostic mention detection model on typed data. Top sub-table shows results on standard english benchmarks, followed by aggregated and mean values. The middle sub-table shows generalization on multi-lingual data, showing transfer to other Latin languages, while bottom rows shows results on LLM-annotated (LLM) datasets. ToMMeR yields high recall on most common NER datasets in a zero-shot setup; while real precision is controlled with LLM judged data. } 
% model hash is  92bf3cebce4887dabb0d77f09f6d2a79c47f6e40f0e7cc36c9af5e50fcc14bc1
\label{tab:ZeroShotResults}
\vspace{-.3cm}
\end{table*}

To evaluate ToMMeR’s ability to detect entity mentions without fine-tuning, we assess its performance across three dimensions: (1) zero-shot transfer to common english NER benchmarks, (2) precision validation using LLM-based judgment, (3) Multi-lingual transfer.
Our findings demonstrate that ToMMeR achieves high recall (92.6\%) with minimal parameters (274K), while LLM-judged precision (92\%) confirms alignment with a broad notion of entities. 

\subsection{Zero-shot Mention Detection}\label{sec:ZeroShotRes}

\paragraph{Datasets and Metrics.}\label{sec:datasets}

We evaluate ToMMeR on 13 NER corpora spanning news, Wikipedia, scientific/biomedical and industrial domains, and multi-genre resources, as well as multi-lingual data. We detail all datasets in Appendix~\ref{sec:datasets-appendix}, and provide a comparison of schemas and annotation methods in Appendix~\ref{app:schema-comparison}. We report recall, precision, and F1 scores on mention detection. We target high recall to capture all potential entity mentions, while lower precision is expected as ToMMeR detects mentions beyond standard benchmark types (Precision and Recall \textcolor{gray}{in gray} in \Cref{tab:ZeroShotResults}). 

\paragraph{Results.} In this zero-shot setup, ToMMeR consistently achieves high recall across most of the 13 tested benchmarks (See \Cref{tab:ZeroShotResults} upper part and average), demonstrating strong coverage of entity mentions and general alignment with the notion of entity captured in these datasets (up to 98.6\% recall on MultiNERD). Precision is as expected lower on gold data due to non-nested and limited scope of annotated entities. 
One exception is on ACE 2005 (40\% recall vs. 88.2\% in average), due to a distinct notion of entity mention learned on Pile-NER. ACE include both determiners (e.g., “the president”), which differs from the patterns ToMMeR learned (e.g. “president”) and pronouns (for coreference resolution, not annotated in Pile-NER), which ToMMeR didn't learn to detect\footnote{A comparison of those schemas is provided in appendix, fig.~\labelcref{fig:SchemaComparison}, as well as random ACE samples with ToMMeR's predictions in App~\labelcref{fig:inferenceACE}. More details in App.~\ref{sec:datasets-appendix} and \ref{app:schema-comparison}.}.
ToMMeR also transfers well on Latin languages, achieving 86\% average recall on French, Spanish, and German when evaluated on WikiANN. Performance however drops on Chinese (41\% recall), likely due to significant differences in tokenization and syntax encoding in the English-centric Llama backbone.

\paragraph{Flat (Non-Nested nor Overlapping) Decoding}\label{sec:decoding}
By design, ToMMeR can predict any \textit{continuous} span, which naturally produces many nested entities under our criteria. Since most applications require flat annotations, we can choose to post-process the predictions to obtain a non-overlapping segmentation of the text. Additionally, by adjusting the decision threshold, ToMMeR offers a flexible trade-off between precision and recall, depending on the application’s needs. 
To evaluate flat (non-nested nor overlapping) segmentation, we implement a \textit{greedy decoding algorithm} that iteratively select the highest scored span that does not overlap with previously chosen ones.
Constraining the model to produce non-overlapping mentions on flat NER datasets further improves precision, increasing the average from roughly 30\% to 40\%. (See greedy decoding columns in \Cref{tab:ZeroShotResults}).

\subsection{Evaluating precision -- LLM as a Judge}\label{sec:LLMasJudgeMethod}

To better evaluate ToMMeR precision, we must determine whether predicted spans that are not labeled in the dataset are false positives.
We use an \textit{LLM-as-a-judge} evaluation to create two LLM-judged datasets, and cautiously validate them  with human judgments to address circularity risks and control biases.

\paragraph{LLM Annotation.}
%\noindent\textbf{LLM Annotation.}
We consider MultiNERD (commonsense) and GENIA (domain-specific) benchmarks. We first generate as many candidate mentions as possible using the ToMMeR model with highest recall on validation data, ensuring we capture the broadest possible set of entity spans. Then, given a span predicted by ToMMeR, we prompt \texttt{gpt-4.1-mini} \cite{openaiAPI} to assess whether it qualifies the predicted span as an entity according to the Wikipedia definition \cite{wikipediaEntity2025}. The complete prompt is provided in the appendix \Cref{fig:LLMasJudgePrompt}. 
We sample 10,000 spans, providing sufficient coverage to produce reliable precision estimates while limiting inference costs.
After removing spans deemed invalid by the LLM judge, we obtain curated datasets containing a high number of nested entity mentions. This approach allows us to capture the precision of our model beyond the limitations of existing benchmark annotations, including nested and otherwise unannotated mentions and providing a more comprehensive and realistic measure of performance. On these curated datasets, {ToMMeR} reaches a precision of 92\% (two LLM rows Table~\ref{tab:ZeroShotResults}), demonstrating that while the model captures nearly all mentions, it rarely produces spurious predictions.

\paragraph{Human validation of LLM judgments.}

We first verify that the entities accepted by the LLM cover the gold-annotated ones: This is the case in more than 99\% of the cases.
We then quantify the positive bias~\cite{thakur-etal-2025-judging} of our judge, which potentially overestimate precision. 
We collected human annotations on predicted spans sampled from LLM-judged data across GENIA (600 spans) and MultiNERD (1,200 spans). Annotations were performed by 5 CS researchers following identical instructions as the LLM judge, with strict blinding to both LLM predictions and other annotators' judgments (full protocol in Appendix \labelcref{app:LLMasJudge}). 
For MultiNERD, human-LLM agreement reaches 91.5\% ($\kappa=0.449$), indicating substantial consensus on common-sense entities. For GENIA, agreement drops to 78.5\% ($\kappa=0.239$), reflecting genuine ambiguity in technical biomedical entities rather than simple judge failure—both non-expert human annotators and the LLM struggle with domain-specific boundaries.

To obtain unbiased precision estimates, we apply the Prediction-Powered Inference (PPI) framework \cite{angelopoulosPredictionPoweredInference2023}. It uses a small human-labeled sample to adjust predictions from a larger LLM-labeled set: providing a scalable, cost-effective method for precision estimation.
This yields a corrected precision of $90.8\pm 2\%$ for MultiNERD and $87.1\pm 5\%$ for GENIA with 95\% confidence (instead of raw LLM-judged precisions of 92.8 and 90.3 respectively), substantially exceeding the low precision using the gold-annotation from standard benchmarks (23.2\%). This confirms that most ToMMeR predictions rejected by benchmarks are genuine entities outside the annotation scope rather than false positives.
Details, including Cohen's $\kappa$ scores, are reported in \cref{tab:agreement_kappa}. More details are reported in Appendix \ref{app:LLMasJudge}.

\begin{table}[t]
\centering
\resizebox{0.9\linewidth}{!}{%
\begin{tabular}{rccc}
\toprule
\textbf{Data} & \textbf{MultiNERD} & \textbf{GENIA} & \textbf{Aggregated} \\
\midrule
Judged Precision    & 92.8              & 92.3           & \textbf{92.55} \\
Human Agreement     & 91.5 \%           & 78.5 \%        & \textbf{87.17 \%} \\
Support             & 1200              & 600            & \textbf{1800} \\
Cohen's $\kappa$    & 0.449             & 0.239          & \textbf{0.359} \\
Bias Correction     & -2.00\%           & -5.17\%        & \textbf{-3.06\%} \\
Rectified Precision & 90.80             & 87.13          & \textbf{89.49} \\
95\% Lower Bound    & 89.42             & 84.03          & \textbf{88.11} \\
\bottomrule
\end{tabular}
\vspace{-5mm}
}
\vspace{-2pt}\caption{Estimating ToMMeR real precision with the PPI framework: using an LLM-as-a-judge, controlled with human annotation. While \texttt{gpt-4.1-mini} does exhibit positive bias, human validation confirms that it can be used as a reliable evaluator. More details in App.~\labelcref{app:LLMasJudge}.}
\label{tab:agreement_kappa}
\vspace{-.7cm}
\end{table}

This approach ultimately captures ToMMeR's performance beyond incomplete benchmark annotations. Even if the GENIA results show higher uncertainty due to domain complexity and annotator expertise limitations, results are consistent across both datasets: high agreement between human and LLM judge, modest positive bias, and strong PPI-corrected precision.

\subsection{Deeper Analysis}\label{subsec:Deep}
\begin{figure}[h]
    \vspace{-.3cm}
    \hspace{-.65cm}
    \includegraphics[scale=.68]{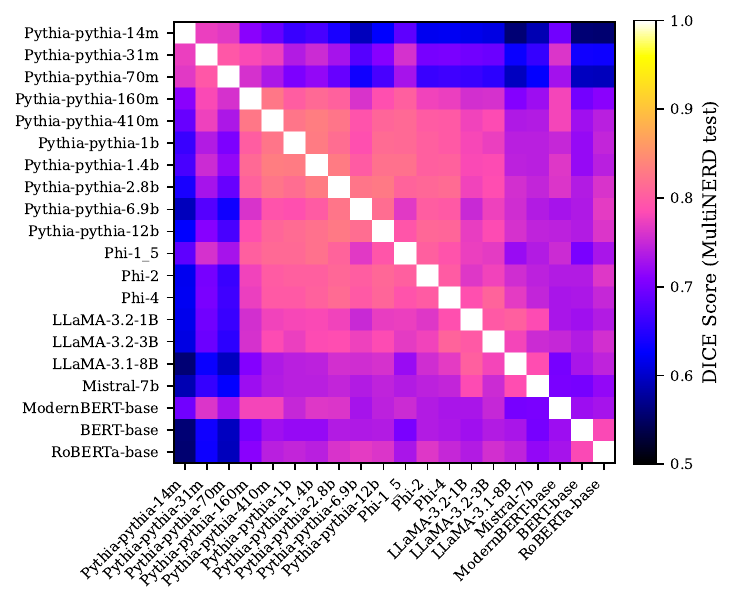}
    \caption{ DICE score between inference of ToMMeR trained on various LLMs on MultiNERD test (Results for GENIA are similar, see Appendix~\labelcref{app:AdditionnalFigs}, fig.~\ref{fig:dice_Models_GENIA}).}
    \label{fig:dice_Models_MultiNERD}
    \vspace{-.5cm}
\end{figure}

\begin{figure*}[t]
\centering
    \includegraphics[scale=.62]{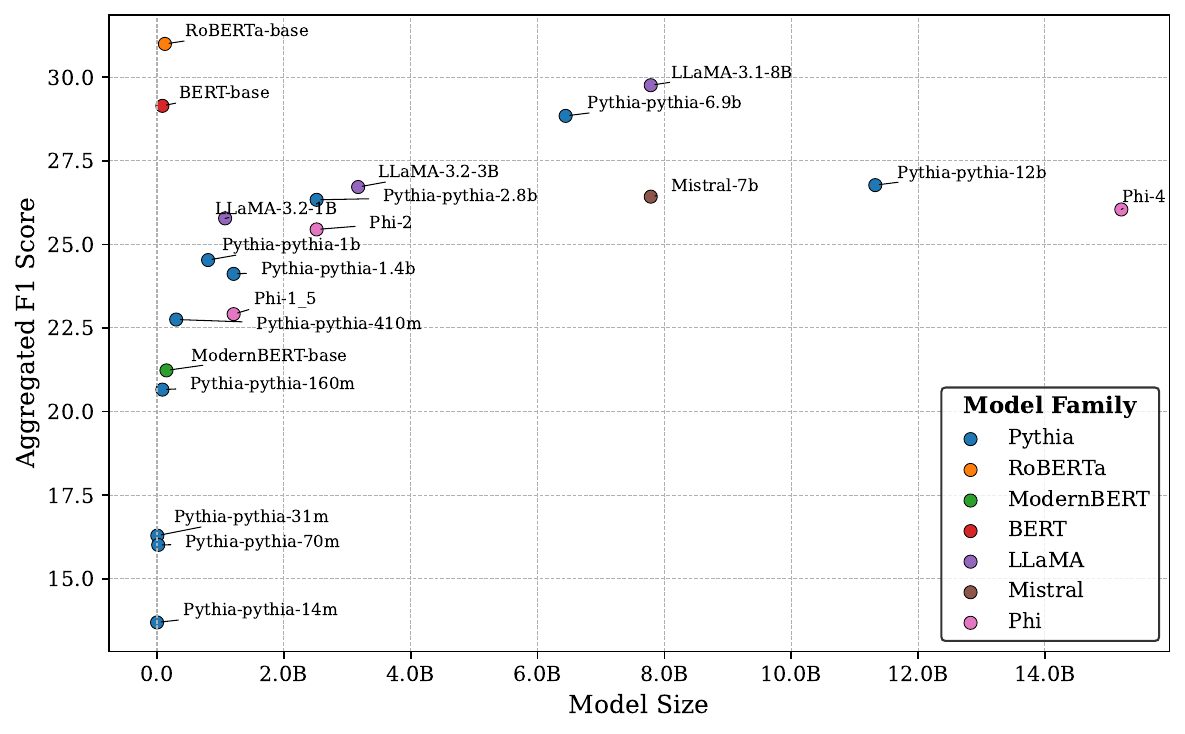}
    \vspace{-0.35cm}
    \caption{F1 Score of ToMMeR Models —aggregated on the 13 benchmarks considered in this work- versus number of parameters of LLM backbone. We also plot the precision versus recall for all those models in appendix \Cref{fig:AllModels_Prec_Recall}.}
    \label{fig:AllModels}
    %\vspace{-.2cm}
\end{figure*}

\paragraph{LLMs share a common Notion of Entity Mention.}
If two models capture the same underlying notion of entity, their predicted entity mention sets should exhibit a high degree of overlap. We measure this similarity by computing the Sørensen–Dice coefficient~\cite{diceMeasuresAmountEcologic1945} for mention predictions across all tested LLMs (See \Cref{fig:dice_Models_MultiNERD}). We consider LLMs ranging from 14M to 15B parameters, providing a broad basis to study the effect of scale while keeping resource use manageable. For decoder-only models, we use the Pythia family, which offers a controlled size sweep for systematic scaling studies. We also include LLaMA 3 models (1B, 3B, and 8B) and \textsc{Mistral-7b}, representing recent state-of-the-art open LLMs. We further include the \textsc{Phi} family --trained with textbook-style synthetic data-- (see in appendix, \Cref{table:modelParams} for details about all the models considered).
To test transferability beyond decoder-only settings, we additionally evaluate encoder-only architectures, including BERT and RoBERTa as established NER baselines, and ModernBERT as a more recent encoder.
We observe that BERT family offers the best model size--F1 trade-off when considering the aggregated on the 13 benchmarks.

Auto-regressive backbones (excluding small Pythia models with <160M p.) reach Dice  >75\%, indicating strong agreement in their learned notion of mentions. This suggests that diverse LLMs—despite differences in scale or training data—develop a convergent and architecture-agnostic representation of entity boundaries.
Encoder-only models predictions slightly differ from decoder-based architectures (Dice drop of $\sim10$\%). From the results, we hypothesize that they predict less nested entities due to their bidirectional attention mechanism, which may suppress overlapping spans in favor of flat, non-hierarchical segmentation. More results  in Appendix \Cref{fig:AllModels_Prec_Recall}.

\vspace{-5pt}
\paragraph{Mention detection capabilities through layers.}\label{sec:layer_xp}

To localize where mention detection capability emerges within the network, we train a \hyperref[sec:ToM]{ToMMeR} model on the hidden representations extracted from each layer of the backbone LLM (here \textsc{LLaMA-3.2-1B}). Maximum performance is nearly reached after the first layer of the transformer (\Cref{fig:Alllayers}). This suggests that mention-detection signals are established very early in the model’s computation and remain largely stable across intermediate layers.

\begin{figure}[h]
    \centering
    % This file was created with tikzplotlib v0.10.1.
\begin{tikzpicture}
\tikzstyle{every node}=[font=\small]

\definecolor{coral25111684}{RGB}{251,116,84}
\definecolor{darkgray176}{RGB}{176,176,176}
\definecolor{darkslateblue3654149}{RGB}{36,54,149}
\definecolor{firebrick2072831}{RGB}{207,28,31}
\definecolor{lightgray204}{RGB}{204,204,204}
\definecolor{lightgreen151214184}{RGB}{151,214,184}
\definecolor{lightseagreen36152192}{RGB}{36,152,192}
\definecolor{mediumturquoise81188193}{RGB}{81,188,193}
\definecolor{palegoldenrod214239178}{RGB}{214,239,178}
\definecolor{peachpuff252201180}{RGB}{252,201,180}
\definecolor{steelblue33100171}{RGB}{33,100,171}

\begin{axis}[
%legend cell align={left},
legend style={
  fill opacity=0.6,
  draw opacity=1,
  text opacity=1,
  at={(0.01,0.01)},
  anchor=south west,
  draw=lightgray204,
  legend columns=3
},
width = 7.5cm,    % Set the desired width
height= 4.9cm,    % Set the desired heigh
tick align=outside,
tick pos=left,
x grid style={lightgray204},
xlabel={layer of \textsc{Llama-3.2-1B}},
xmajorgrids,
xmin=-0.5, xmax=16,
xtick style={color=black},
y grid style={darkgray176},
ylabel={Score},
ymin=0.09, ymax=1,
ytick={0.1,0.3,...,0.9},
yticklabels={0.1,0.3,0.5,0.7,0.9},  % force label text
ymode=log,
ymajorgrids,
ytick style={color=black}
]

\addplot [dashed, firebrick2072831, opacity=1, solid, mark=*, mark size=1, mark options={solid}]
table {%
0 0.934887602264685
1 0.9380093248407646
2 0.9429984099079972
3 0.9417167156404812
4 0.939432410757254
5 0.9341497704175512
6 0.9262095674451523
7 0.9417628580325549
8 0.9368416520877565
9 0.9430186089171974
10 0.9333592583156404
11 0.9309936713375796
12 0.9365647085633404
13 0.9308987881104034
14 0.9299156795470629
15 0.9473782125973107
};
\addlegendentry{Recall}

\addplot [dashed, darkslateblue3654149, opacity=1, solid, mark=*, mark size=1, mark options={solid}]
table {%
0 0.1828080141542817
1 0.18616188874734607
2 0.18029805208775654
3 0.18920921698513798
4 0.1989283323425336
5 0.21771293078556264
6 0.2319050819532909
7 0.188590425477707
8 0.20187658259023353
9 0.18995565520169852
10 0.20973233347487616
11 0.21120168775654638
12 0.18859924529370137
13 0.19609613588110403
14 0.1978628523708422
15 0.11839315612172682
};
\addlegendentry{Precision}

\addplot [dashed, steelblue33100171, opacity=1, solid, mark=*, mark size=1, mark options={solid}]
table {%
0 0.3058166168263856
1 0.3106672461708486
2 0.3027175499621691
3 0.3151072537075447
4 0.328331285057473
5 0.3531262607910475
6 0.37093513281411206
7 0.314251235762166
8 0.33217416809900563
9 0.3162149810411972
10 0.3425020647913107
11 0.34429737979868047
12 0.3139727265493325
13 0.32395115782478595
14 0.3262976969541137
15 0.2104824729250627
};
\addlegendentry{F1}

\end{axis}

\end{tikzpicture}
    \vspace{-.9cm}
    \caption{Layer-wise Performance on \textsc{Llama3.2-1B}. Recall, precision and F1 Score of ToMMeR probing representations across the $16$ layers ($0$-$15$) of \textsc{Llama3.2-1B}. Performance is nearly optimal from layer $0$ onward.}
    \label{fig:Alllayers}
    \vspace{-.5cm}
\end{figure}

\begin{table*}[t]
\centering
\resizebox{0.92\textwidth}{!}{%
\begin{tabular}{lccccccc}
\toprule
Model & Backbone & \#Trained Params & CoNLL 2003 & GENIA & MultiNERD & OntoNotes & ncbi \\
\midrule
GLiNER (\citeyear{zaratianaGLiNERGeneralistModel2023})
    & deBERTa-v3 & 209M & \textbf{88.7} & \textbf{78.9} & \textbf{93.8} & \textbf{89.0} & \textbf{87.8} \\
EMBER (\citeyear{popovicEmbeddedNamedEntity2024})
    & GPT2-xl (0–47) & 11.5M & 85.1 & – & – & 79.3 & – \\
\midrule
\multirow{5}{*}{\makecell[l]{ToMMeR (ours)\\ + span embed ($\ell 11$)}} 
    & LLaMA-3.2-1B ($\ell6$) & 7.6M & 84.8 & 66.5 & 92.2 & 80.4 & 78.1 \\
    & LLaMA-3.2-3B ($\ell5$) & 7.7M & 86.8 & 69.3 & \textbf{93.3} & 81.7 & \textbf{82.1} \\
    & LLaMA-3.1-8B ($\ell5$) & 7.9M & 85.0 & \textbf{70.1} & 92.4 & 80.0 & 80.8 \\\cmidrule(lr){2-8}
    & RoBERTa-base ($\ell5$) & 7.4M & \textbf{87.3} & 67.8 & 92.6 & \textbf{85.4} & 74.8 \\
    & BERT-base ($\ell3$) & 7.4M & 85.0 & 66.5 & 90.4 & 82.1 & 77.3 \\
\bottomrule
\end{tabular}%
}
%\vspace{-3pt}
\caption{Micro-F1 scores when \hyperref[sec:fullNER]{training a classification head on top of a ToMMeR model} to perform full NER. Baselines are reported from \citet{zaratianaGLiNERGeneralistModel2023} and \citet{popovicEmbeddedNamedEntity2024}. We also report in parenthesis the transformer layers probed to perform mention detection.}
\label{tab:ft_classification}
\vspace{-.3cm}
\end{table*}

We nonetheless observe a noticeable loss at the final layer, where the model likely discards these signals to prioritize next-token prediction. We also show the Dice similarity scores across layers in appendix (\cref{fig:dice_Layers_MultiNERD}), showing that entity-related signals are not only learned early in the network but also remain largely consistent across layers. 
%Together, these results suggest that mention-tracking information is stably maintained across the model’s depth.

%\section{Extension to Full NER via Span Classification}\label{sec:fullNER}
\section{Full NER via Span Classification}\label{sec:fullNER}
\vspace{-5pt}
To further assess the utility of ToMMeR as a generalist mention detection framework, we conduct an extrinsic evaluation by applying it within a complete NER pipeline that includes entity typing.

\paragraph{Approach.} For each NER benchmark, we first finetune ToMMeR to align with its notion of entity mention, ensuring high recall on the benchmark. Then, for each predicted mention, we compute a $d$-dimensional span embedding used for classification: Following \citet{zaratianaGLiNERGeneralistModel2023}, we compute this embedding by passing the concatenation of the spans first and last tokens representations through a two-layer perceptron (MLP) of hidden dimension 1024. We train this representation layer on the benchmark schema using standard Cross-Entropy loss, an additional \texttt{0} (not-entity) label is added to filter mentions detected by ToMMeR but not in the benchmarks schema.
This lightweight architecture (7.4M params) enables us to adapt our ToMMeR model (using \textsc{Llama-3.2-1B} representations) to CoNLL in less than 20 minutes on a V100-32Gb.

\paragraph{Results.} We benchmark ToMMeR  using a range of backbone LLMs, and report the results in \Cref{tab:ft_classification}. Despite being attached to autoregressive models that cannot exploit right-side context, ToMMeR achieves near SOTA performance on multiple datasets. On OntoNotes, for instance, the \textsc{LLaMA-3.2-1B}–based ToMMeR reaches an F1 score of 80.4\%, even though its initial zero-shot mention detection recall was only 73\% (\Cref{tab:ZeroShotResults}). This shows ToMMeR can effectively adapt to a dataset-specific notion of entity mention during fine-tuning.
In comparison to EMBER \cite{popovicEmbeddedNamedEntity2024} --trained end-to-end on the NER task and identifying mention detection as the main performance bottleneck-- ToMMeR achieves comparable performance. This further supports our approach of probing existing mention detection capabilities in LLMs and demonstrates their utility for downstream NER tasks. We extend our analysis to encoder-only architectures, which are both more parameter-efficient and well-suited for NER due to their bidirectional attention mechanisms. Interestingly, these models do not consistently outperform their autoregressive counterparts under the ToMMeR + typing setup, suggesting that entity-level representations may emerge differently across architectural families. Moreover, larger LLMs may also encode a broader spectrum of entity types.

\section{Conclusion}
\label{sec:conclusion}
\vspace{-6pt}
We showed that mention detection capabilities —crucial for information extraction— can be efficiently probed from early LLM layers using less than 300K parameters. 
Our work provides both practical and conceptual contributions: practically, it offers a lightweight, transferable method for high-coverage mention detection that can be plugged into any LLM; conceptually, it provides evidence that LLMs develop structured mention representations in their early layers that can be recovered through simple probing mechanisms.
ToMMeR achieves 93\% recall across 13 diverse NER benchmarks, transfers well zero-shot to other latin languages, while maintaining an estimated 90\% precision using a single partial forward pass, enabling real-time streaming deployment with minimal overhead (Efficiency analysis in Appendix \labelcref{app:Efficiency}). 
Across models from 14M to 15B parameters, we find that diverse architectures converge on a shared notion of entity mention, producing highly consistent spans. This suggests mention tracking emerges as a byproduct of language modeling rather than as an artifact of a particular architecture.
When extended with a typing head, ToMMeR achieves competitive full NER performance despite using auto-regressive models lacking right-side context. Unlike costly prompt-based extraction methods, ToMMeR reduces costs by orders of magnitude ($42\times$ faster than prompting) while maintaining flexibility to adapt to any downstream schema or knowledge base. ToMMeR's schema-agnostic design integrates naturally into modern RAG/IE pipelines as a first-stage mention filter, enabling entity-aware document chunking, or providing spans for downstream typing and linking.
\section*{Limitations}

\paragraph{Lack Ground Truth for Untyped Mention Detection.}

The fundamental challenge in evaluating our approach is the lack of established ground truth for untyped mention detection. While typed NER benchmarks provide gold annotations, they only label entities from specific ontologies (e.g., person, location, organization), making it unclear whether unlabeled spans are true negatives or simply out-of-scope entities. We address this by using LLM-based judgment aligned with Wikipedia's entity definition, but this introduces its own limitations: (1) \texttt{gpt-4.1-mini} is slightly biased towards positive judgements (see \Cref{app:LLMasJudge}) (2) the Wikipedia definition is broad and potentially ambiguous in edge cases. While human validation shows reasonable agreement with the LLM judge (87.7\% agreement on 1800 annotated mentions ), comprehensive human annotation would be needed to definitively establish ToMMeR precision. Our PPI analysis however bounds the true underlying precision to above 89\%.

\paragraph{Architectural Constraints.}
Our focus on continuous spans excludes discontinuous entities (e.g., "New York" and "City" separated by other tokens), which appear in some linguistic phenomena and specialized domains. Additionally, our auto-regressive models lack access to right-side context, potentially missing boundary information that bidirectional encoders naturally capture. While Section~\ref{sec:fullNER} shows that competitive full NER performance is still achievable, this architectural limitation may impact mention detection quality compared to encoder-only approaches.

\paragraph{Dataset and Training Limitations.}
Training on Pile-NER, despite its broad coverage, inherits the biases and gaps of GPT-3.5's annotations from \citeyear{Zhou2023UniversalNER}. The distillation strategy, while mitigating some incompleteness, risks amplifying systematic biases present in the initial annotations.

\paragraph{Schema Generalization.}
While we evaluate across 13 benchmarks (spanning news, biomedical and general domains) and obtain strong average recall. The ACE 2005 results (42\% recall) demonstrate that our models struggle with annotation conventions that differ substantially from Pile-NER, such as including determiners in entity spans. This suggests that while the learned notion of mentions generalizes well across most domains, it may not align with all possible annotation schemes, although we also show that ToMMeR can easily further be tuned to fit to such schemes.

\section*{Acknowledgements}
The authors acknowledge the ANR – FRANCE (French National Research Agency) for its financial support of the GUIDANCE project n°ANR-23-IAS1-0003 as well as the Chaire Multi-Modal/LLM ANR Cluster IA ANR-23-IACL-0007. This work was granted access to the HPC resources of IDRIS under the allocation 2024-AD011015440R1 made by GENCI. The authors also gratefully acknowledge the support of the Centre National de la Recherche Scientifique (CNRS) through a research delegation awarded to J. Mothe.

\bibliography{Recherche,related,datasets}
%\section*{Acknowledgments}

\appendix
\section{Additional Examples and Figures}\label{app:AdditionnalFigs}

\subsection{Figures}

\begin{figure}[h]
    \centering
    %\hspace{-.6cm} % sinon les marges ACL passent pas...
    \includegraphics[scale=.8]{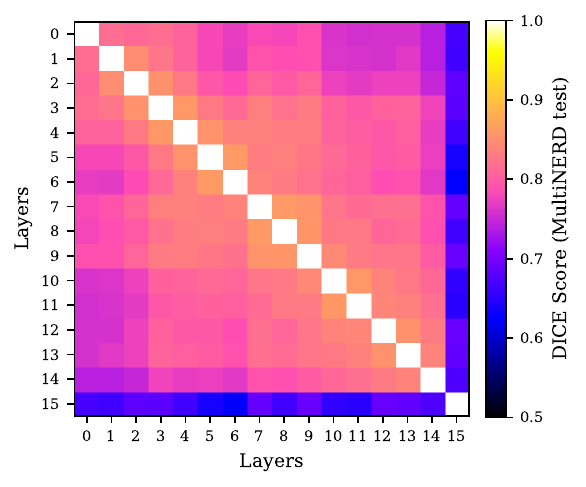}
    \vspace{-0.5cm}
    \caption{ DICE score between the sets of entities inferred on the MultiNERD (test), for ToMMeR models probing each layer of \textsc{Llama3.2-1B}. Results for GENIA are similar, and available in appendix, \Cref{fig:dice_Layers_GENIA}.}
    \label{fig:dice_Layers_MultiNERD}
    \vspace{-0.5cm}
\end{figure}

\begin{figure}[h]
    \centering
    \hspace{-.6cm} % sinon les marges ACL passent pas...
    \includegraphics[scale=.8]{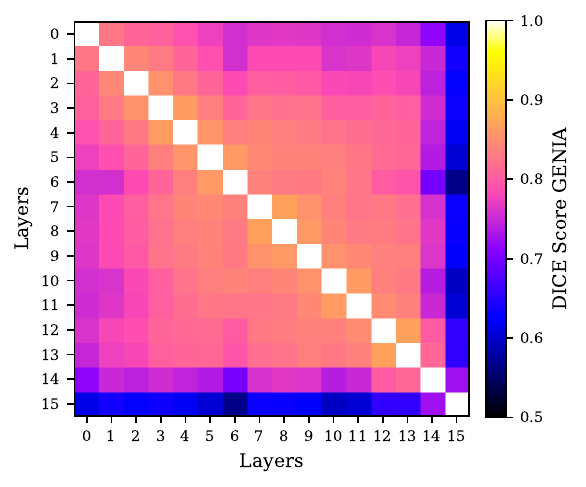}
    \caption{ DICE score between inference for models using all possible layers of \textsc{Llama3.2-1B}, on the full GENIA dataset, results are similar to those in \Cref{fig:dice_Layers_MultiNERD}.}
    \label{fig:dice_Layers_GENIA}
\end{figure}

\begin{figure}[h]
    \includegraphics[scale=.65]{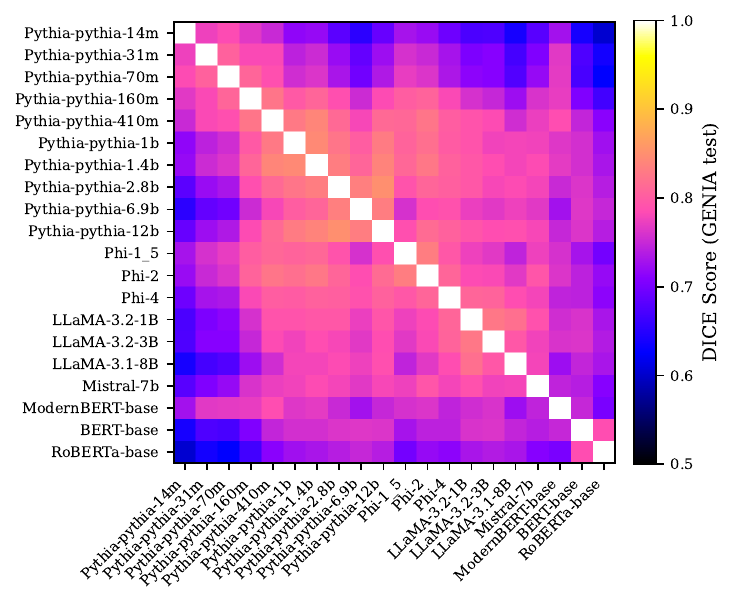}
    \caption{DICE score between inference of ToMMeR trained on various existing LLMs on GENIA test.}
    \label{fig:dice_Models_GENIA}
\end{figure}

Figure~\ref{fig:dice_Layers_MultiNERD} presents the DICE score between the sets of entities inferred on the MultiNERD (test split) for ToMMeR using all possible layers of LLAMA3.2-1B. Figure~\ref{fig:dice_Layers_GENIA} presents the results for GENIA. 
Figure~\ref{fig:dice_Models_GENIA} is also the twin figure of Figure~\ref{fig:dice_Models_MultiNERD} in the same Sub-section.

\subsection{Pile-NER statistics}\label{app:PileNERStats}

\begin{figure}
    \centering
    \includegraphics[width=\linewidth]{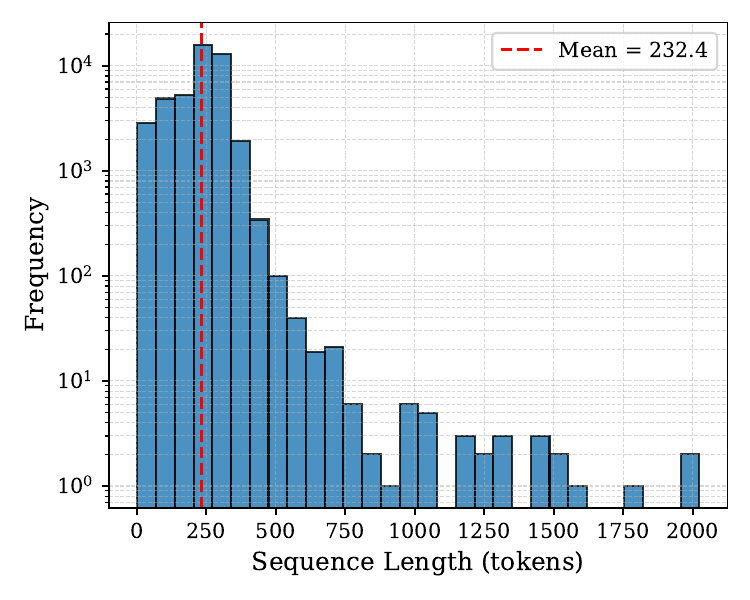}
    \caption{Sample token length distribution in Pile-NER \cite{zhouUniversalNERTargetedDistillation2023} using \textsc{Llama} tokenizer.}
    \label{fig:PileNERSeqLength}
\end{figure}

\begin{figure}
    \centering
    \includegraphics[width=\linewidth]{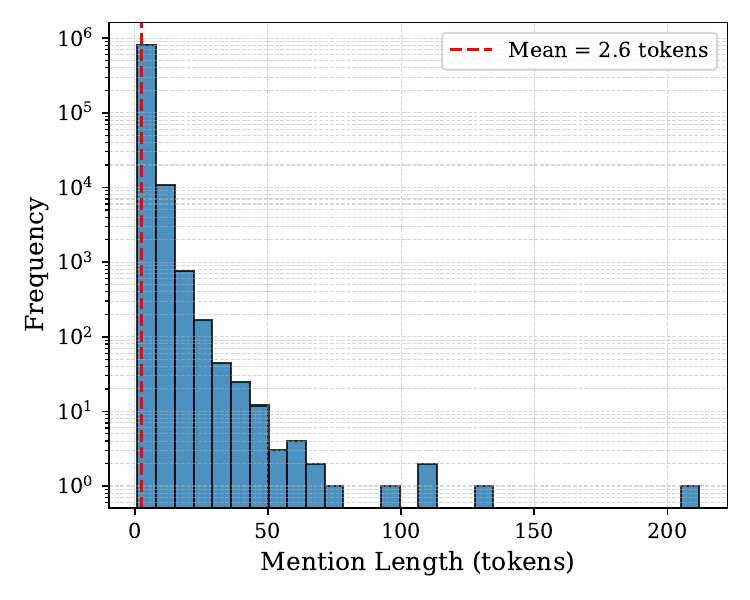}
    \caption{Entity mention length distribution in Pile-NER \cite{zhouUniversalNERTargetedDistillation2023} using \textsc{Llama} tokenizer. We use a sliding window of 25 tokens, which includes 99.8\% of mentions annotated in Pile-NER.}
    \label{fig:PileNEREntityLength}
\end{figure}

Figure~\ref{fig:PileNERSeqLength} reports the distribution of texts token-length of in Pile-NER used in ToMMeR entity mention boundary training. While \Cref{fig:PileNEREntityLength} reports the distribution of entity mention lengths.

\subsection{Qualitative examples}
\begin{figure*}[h]
    \centering
    \includegraphics[scale=.8]{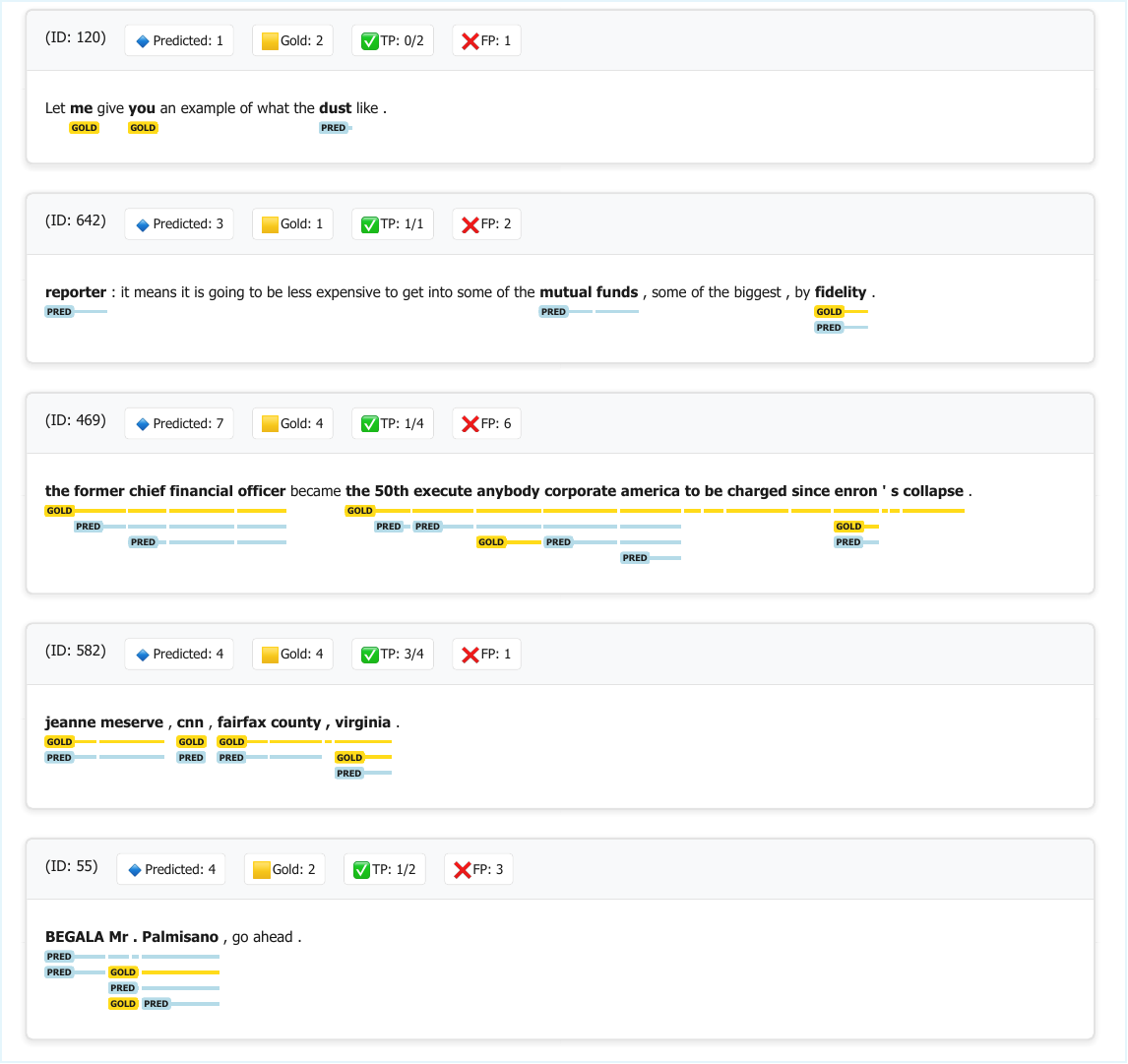}
    \label{fig:inferenceACE}
\caption{Qualitative examples comparing model predictions and gold annotations, randomly sampled from ACE 2005 Dataset where our \hyperref[sec:ToM]{ToMMeR} model has a surprisingly low Zero Shot Recall ( 42\%, Cf \Cref{tab:ZeroShotResults}).}
\end{figure*}

We provide in  \Cref{fig:inferenceACE} some qualitative examples of our models prediction on the ACE benchmark, where ToMMeR has lower zero-shot recall as discussed in \Cref{sec:ZeroShotRes}, suggesting that ToMMeR learned a slightly different notion of entity mention from Pile-NER.

\section{Complexity and efficiency analysis}\label{app:Efficiency}

\paragraph{Complexity}
Unlike prompting methods that require auto-regressive generation ($O(N\cdot M)$ complexity where $M$ is output length), ToMMeR classifies spans in a single parallelizable pass ($O(N)$). It is computationally equivalent to one additional attention head.
On a standalone usage, ToMMeR is faster than one backbone forward pass. As we can cut the computation to at a lower layer (e.g layer 6 / 16 for Llama3-1B gives a $\approx2,6\times$ speedup). And even faster when compared with prompting methods using several forward passes for generation.
It furthermore allows for "streaming extraction"—identifying entities as the LLM generates text with almost zero latency penalty (as shown by \citeauthor{popovicEmbeddedNamedEntity2024}). This is can be useful for real-world RAG pipelines where re-prompting the LLM for extraction doubles the cost and latency.

\paragraph{Quantitative Runtime comparison}
To compare with a prompting baseline, we applied the prompt from UniversalNER \cite{Zhou2023UniversalNER} : \texttt{"Given a passage, your task is to extract all entities and identify their entity types. The output should be in a list of tuples of the following format: [("entity 1", "type of entity 1"), ... ]. Passage: {passage}"} with llama3.2-1B.
Running it on the passage \texttt{"Large language models are awesome. While trained on language modelling, they exhibit emergent abilities that make them suitable for a wide range of tasks, including Named Entity Recognition (NER)."} took \textbf{3.21 seconds} (24Bg GPU, using HF transformers, 100max tokens generation), While ToMMeR inference took \textbf{ only 0.077 seconds, which makes a 42x improvement.}

\section{Reproducibility Statement}\label{app:Reproducibility}
For complete reproducibility, we publish both code at \githubrepo{https://github.com/VictorMorand/llm2ner}, containing detailed hyperparameters and experimental pipeline, and trained ToMMeR models \href{https://huggingface.co/models?other=arxiv:2510.19410}{on hugginface}. We also list in this section the most important hyper-parameters used in our main experiments.

\subsection{Layer experiment}\label{app:LayersHyperparams}
\Cref{tab:hyperparams_layers} details the hyperparameters used in the layer experiment described in \Cref{sec:layer_xp}, where we train ToMMeR models at each layer of \textsc{Llama-3.2-1B}, showing that mention detection signals are computed as early as layer $0$ in the transformer.

\begin{table*}[h]
\centering
\caption{Hyperparameter configuration for the layers experiment.}
\label{tab:hyperparams_layers}
\begin{tabular}{rcl}
\toprule
\textbf{Parameter}          & \textbf{Value}  & \textbf{description} \\\midrule
\texttt{model\_name}                  & Llama-3.2-1B &  LLM backbone \\
\texttt{rank}                        & 64         &  rank $r$ of the ToMMeR model \\
\texttt{optimizer}                   & AdamW     &  optimizer to use  \\
\texttt{epochs}                      & 8         &  number of epochs \\
\texttt{batch\_size}                  & 16        &  batch size   \\
\texttt{sliding\_window}             & 25        &  Sliding window   \\
\texttt{lr}                          & 1e-2      &  learning rate    \\
\texttt{patience}                    & 5000      &  patience for lr scheduler    \\
\texttt{accumulation\_steps}          & 1         &  1 for no accumulation    \\
\texttt{grad\_clip}                   & 2.0       &  0 for no clipping    \\
\texttt{val\_metric}                  & "f1"      &  metric to use for validation \\
\texttt{self\_distillation\_phases}    & 1         &  number of self-distillation phases   \\
\texttt{reset\_student\_weights}       & true      &  whether to reset student weights \\
\texttt{sparse\_distill\_loss}         & true      &  whether to use sparse distillation loss  \\
\texttt{teacher\_thr\_prob}            & 0.90      &  teacher threshold probability    \\
\bottomrule
\end{tabular}
\end{table*}

\subsection{Hardware settings}
Training the rank $64$ ToMMeR model using representations from layer $6$ of \textsc{Llama-3.2-1B} takes 4 hours on a NVIDIA-H100-80Gb GPU (though it could be run on smaller GPUs, peak GPU memory being only 6GBs) using  hyperparameters described in \Cref{app:LayersHyperparams}. We also leveraged V100-32Gb for evaluations and inference.

\subsection{Model experiment}\label{app:ModelsHyperparams}
The goal of the model experiment is to compare ToMMeR performance across LLM backbones. To moderate variance, we trained several models, using representations from different layers, we keep the best performing ToMMeR model for each backbone. Hyper-parameters are detailed in \Cref{tab:hyperparams_models}

\begin{table*}[h]
\centering
\caption{Hyperparameter configuration for the model experiment.}
\label{tab:hyperparams_models}
\begin{tabular}{rcl}
\toprule
\textbf{Parameter}          & \textbf{Value}  & \textbf{description} \\\midrule
\texttt{model\_name}                  & Llama-3.2-1B &  LLM backbone \\
\texttt{layer}                        & [1, 3, 5]. &  layers $l$ of the LLM to extract \\
\texttt{rank}                        & 64          &  rank $r$ of the ToM model \\
\texttt{optimizer}                   & AdamW       &  optimizer to use  \\
\texttt{epochs}                      & 8           &  number of epochs \\
\texttt{batch\_size}                  & 16         &  batch size   \\
\texttt{sliding\_window}             & 25          &  Sliding window   \\
\texttt{lr}                          & 1e-2       &  learning rate    \\
\texttt{accumulation\_steps}          & 1         &  1 for no accumulation    \\
\texttt{grad\_clip}                   & 1.0       &  Gradient clipping    \\
\texttt{val\_metric}                  & "f1"      &  metric to use for validation \\
\texttt{self\_distillation\_phases}    & 1         &  number of self-distillation phases   \\
\texttt{reset\_student\_weights}       & true      &  whether to reset student weights \\
\texttt{sparse\_distill\_loss}         & true      &  whether to use sparse distillation loss  \\
\texttt{teacher\_thr\_prob}            & 0.90      &  teacher threshold probability    \\
\bottomrule
\end{tabular}
\end{table*}

\paragraph{Models considered in this work}

\begin{figure*}[h]
\centering
    \includegraphics[scale=.7]{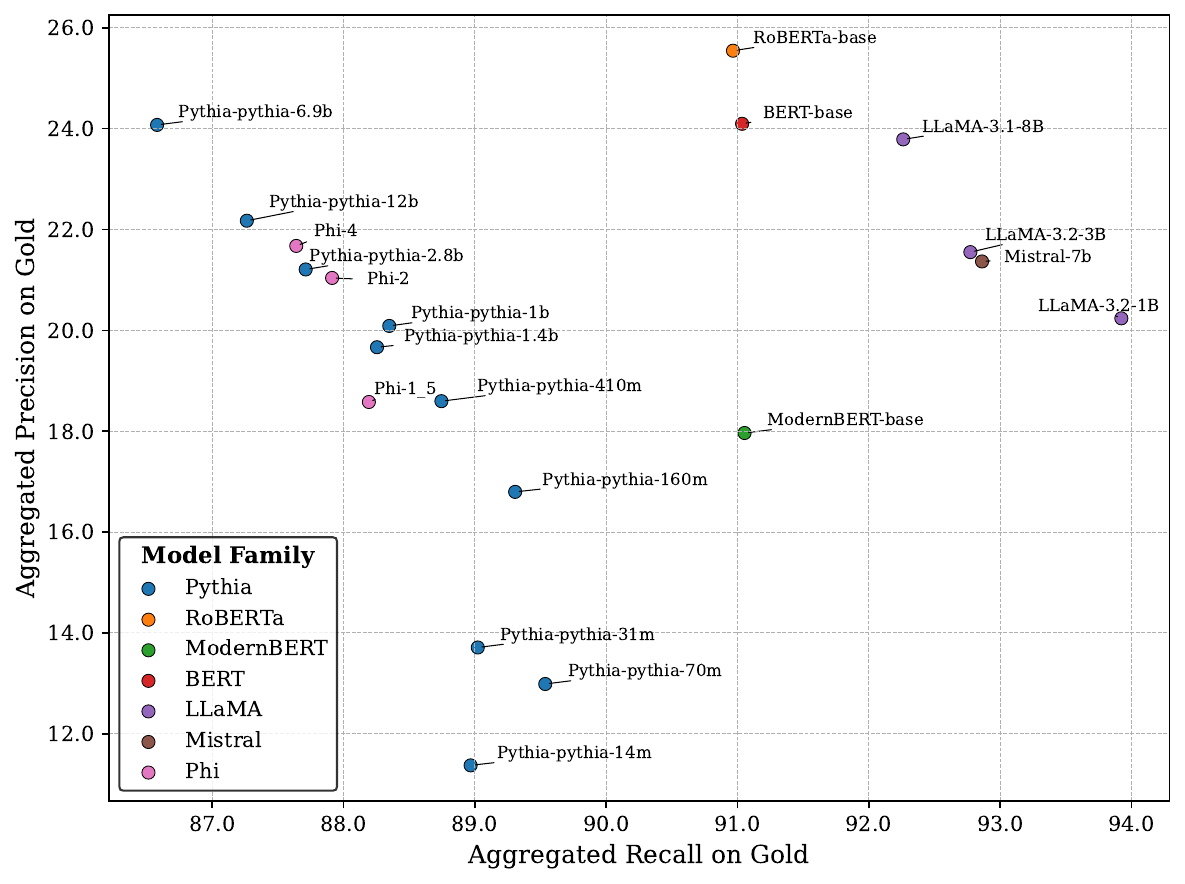}
    \caption{Aggregated precision vs recall of ToMMeR Models versus number of parameters of LLM backbone.}
    \label{fig:AllModels_Prec_Recall}
\end{figure*}

We detail all models architecture parameters in \Cref{table:modelParams}. Please also find in \Cref{fig:AllModels_Prec_Recall} the precision versus recall curve for those models.

\begin{table*}[ht]
\centering
\resizebox{0.8\textwidth}{!}{%
\begin{tabular}{@{}cccccccccc@{}}
\toprule
\textbf{Model name} & \textbf{\# Params} &\textbf{N layers} & \textbf{Model dim} & \textbf{Context length} & \textbf{N Vocab}\\ \toprule
        EleutherAI/pythia-14m & 1.2M & 6 & 128 & 2048 & 50304 \\ 
        EleutherAI/pythia-31m & 4.7M & 6 & 256 & 2048 & 50304 \\ 
        EleutherAI/pythia-70m & 19M & 6 & 512 & 2048 & 50304 \\ 
        EleutherAI/pythia-160m & 85M & 12 & 768 & 2048 & 50304 \\ 
        EleutherAI/pythia-410m & 302M & 24 & 1024 & 2048 & 50304 \\ 
        EleutherAI/pythia-1b & 805M & 16 & 2048 & 2048 & 50304 \\ 
        EleutherAI/pythia-1.4b & 1.2B & 24 & 2048 & 2048 & 50304 \\ 
        EleutherAI/pythia-2.8b & 2.5B & 32 & 2560 & 2048 & 50304 \\ 
        EleutherAI/pythia-6.9b & 6.4B & 32 & 4096 & 2048 & 50432 \\ 
        EleutherAI/pythia-12b & 11B & 36 & 5120 & 2048 & 50688 \\\midrule
        meta-llama/Llama-3.1-8B & 7.8B & 32 & 4096 & 2048 & 128256 \\ 
        meta-llama/Llama-3.2-1B & 1.1B & 16 & 2048 & 2048 & 128256 \\ 
        meta-llama/Llama-3.2-3B & 3.2B & 28 & 3072 & 2048 & 128256 \\\midrule
        mistralai/Mistral-7B-v0.1 & 7.8B & 32 & 4096 & 2048 & 32000 \\\midrule
        microsoft/phi-1\_5 & 1.2B & 24 & 2048 & 2048 & 51200 \\ 
        microsoft/phi-2 & 2.5B & 32 & 2560 & 2048 & 51200 \\ \midrule
        answerdotai/ModernBERT-base & 149M & 22 & 768 & 8192 & 50368 \\ 
        FacebookAI/roberta-base & 125M & 12 & 768 & 512 & ~ \\ 
        google-bert/bert-base-uncased & 85M & 12 & 768 & 512 & 30522 \\ \bottomrule

\end{tabular}%
}%
\caption{Architecture characteristics of the LLMs considered in this work.}
\label{table:modelParams}
\end{table*}

\section{LLM as a Judge}\label{app:LLMasJudge}

\subsection{LLM-as-a-judge}
As described in \Cref{sec:LLMasJudgeMethod}, the goal of this experiment is to have an idea of the real precision of ToMMeR. Given a span predicted by ToMMeR, we prompt \texttt{gpt-4.1-mini} (via the OpenAI API \cite{openaiAPI}) to assess whether it qualifies the predicted span as an entity according to the Wikipedia definition \cite{wikipediaEntity2025}. While it is hard for the model to predict directly the correct true/false tokens when given the context, we find that letting the model generate a small explanation before answering greatly improves accuracy. The complete prompt is provided \Cref{fig:LLMasJudgePrompt} along with an example answer.

The prompt used to judge the inference of our ToMMeR models can be found \Cref{fig:LLMasJudgePrompt}. We now give more details on the human validation study.

\subsection{Human validation}

\paragraph{Human Annotation Protocol.}
We recruited five annotators from among the authors and colleagues—all informatics researchers, though not all NLP specialists. To ensure consistency, annotators followed identical instructions to the LLM judge, augmented with clarified examples from author-annotated data.  The annotation process followed strict blinding protocols: annotators had no access to (1) other annotators' judgments or (2) the LLM's predictions, preventing potential biases. Complete inter-annotator agreement metrics and Cohen's $\kappa$ statistics are presented in \cref{tab:agreement_kappa_full}, including per-dataset disagreements breakdowns.

\begin{table*}[ht]
\centering
\resizebox{0.8\textwidth}{!}{%
\begin{tabular}{
    l
    >{\centering\arraybackslash}p{2.5cm}
    >{\centering\arraybackslash}p{2.5cm}
    >{\centering\arraybackslash}p{2cm}
    >{\centering\arraybackslash}p{2cm}
    >{\centering\arraybackslash}p{2cm}
}
\toprule
\textbf{Dataset} &
\textbf{Agreement Rate (\%)} &
\textbf{Correct / Total} &
\textbf{Cohen's $\kappa$} &
\makecell[l]{Human=True \\ LLM=False} &
\makecell[l]{Human=False \\ LLM=True} \\
\midrule
GENIA & 78.50 & 471 / 600 & 0.239 & 49 & 80 \\
MultiNERD & 91.50 & 1098 / 1200 & 0.449 & 39 & 63 \\
\cmidrule(lr){1-6}
\textbf{Aggregated} & \textbf{87.17} & \textbf{1569 / 1800} & \textbf{0.359} & \textbf{88} & \textbf{143} \\
\bottomrule
\end{tabular}
}
\vspace{2mm}
\caption{Agreement rates and disagreement analysis between human annotators and the LLM across datasets. While the model exhibits a tendency toward slight over-prediction, it can be regarded as a reasonably reliable evaluator.}
\label{tab:agreement_kappa_full}
\end{table*}

% FULL PROMPT
\lstset{basicstyle=\ttfamily\footnotesize, 
  keywordstyle=\color{blue}\bfseries,
  commentstyle=\color{gray}\itshape,
  backgroundcolor=\color{gray!10},
  frame=single,
  breaklines=true,
}

\begin{figure*}[t] % 'p' forces a float page
\begin{lstlisting}
[System:]
You are an expert in entity mention annotation.
A mention is defined as : "something that exists as itself. It does not need  to be of material existence."
In particular, abstractions and legal fictions are usually regarded as entities.  In general, there is also no presumption that an entity is animate, or present. It may  refer to animals; natural features such as mountains; inanimate objects such as tables;  numbers or sets as symbols written on a paper; human contrivances such as laws, corporations and academic disciplines; or supernatural beings such as gods and spirits."

## Instructions
- For each text span provided in [[...]], quickly determine if it is a valid mention as  defined above, regardless of its type, length, or style, but ensuring it is not a fragment.
- Briefly explain in one concise sentence whether the span fits the definition. Then answer with a clear "yes" or "no".

[User:]
"...here that she met her future [[second husband]] , Gottfried Lessing ..."
________________________________________
[Answer: ]
"The span "second husband" refers to a specific person as a distinct entity, fitting the definition of a mention. Yes"
Parsed answer: TRUE
\end{lstlisting}
\caption{Prompt used to annotate our NER inference data using OpenAI API (\texttt{gpt-4.1-mini}, accessed September 2025) \cite{openaiAPI}. we use the definition of entity from Wikipedia \cite{wikipediaEntity2025}, more details in \Cref{app:LLMasJudge}.}
\label{fig:LLMasJudgePrompt}
\end{figure*}

\section{Architecture variants}\label{app:OtherArchis}
Along with the \hyperref[sec:ToM]{ToMMeR} architecture for computing our matching scores $m_{ij}$, we explored other variants.

\paragraph{Normalization function}
In Equation~\ref{eq:TM_qk}, we normalize the dot product between query and key projections using cosine similarity (via $\ell_2$ normalization), which naturally bounds matching scores to $[0,1]$. We also experimented with alternative activation functions to normalize the raw dot products: $\arctan(\cdot)$ and $\log\sigma(\cdot)$ (log-sigmoid). However, both alternatives yielded inferior performance in preliminary experiments. Cosine similarity proved most effective, likely because: (1) it provides scale-invariant matching that is robust to variations in representation magnitudes across layers and models, and (2) its geometric interpretation as angular similarity aligns well with the binding hypothesis—tokens that belong to the same mention should have aligned representations in the learned query-key subspace. All reported results use cosine similarity unless otherwise specified.

\paragraph{Linear Transformation of Queries and Keys (LTQK)}\label{sec:LTQK}

Since the transformer model already computes query and key vectors $\{q_i^h,k_i^h\}_{h,i} \in \mathbb R^{d_h}$ for each attention head $h \in [1,N_h]$ with dimension $d_h$, we can use them directly instead of the representations for the residual stream to compute new queries and keys for our model. The matching score $m_{ij}$ is then computed as:

\begin{equation}\label{eq:TM_qlqk}
    m_{ij} = \sum_{h=1}^{N_h} \ \cos\left( W_Q^h \ q_i^h \ | \ \underbrace{W_K^h \ k_j^h}_{\in \mathbb R^r} \right) \in \mathbb [0,1]
\end{equation}

With ${W_Q^h, W_K^h}_h \in \mathbb{R}^{r \times d_h}$ as the query and key matrices, the model’s queries and keys are already in a lower-dimensional space, making computations lighter and keeping the number of trainable parameters low. For example, using rank $r = 16$ results in only $68$K parameters.

\paragraph{Using existing Attention scores (LCAttn)}\label{sec:LCAttn}
Even further, we can directly leverage the model’s attention scores $a_{ij}^h = \langle q_i^h \ |\  k_j^h \rangle $. 

\begin{equation}\label{eq:TM_CLattn}
    m_{ij} = \log \sigma \left( \sum_{l=0, h=1}^{N_L, N_h} w_h^l \  a_{ij}^h \right) \in \mathbb R
\end{equation}

This approach treats attention as a natural proxy for token binding, and have closely been explored in \citep{popovicEmbeddedNamedEntity2024} assuming that some heads are already specialized for mention detection and that a linear combination of their scores can recover this capability from the model.

\paragraph{Comparing Architectures}

Aggregated results for \textsc{LLaMA-3.2-1B} (\Cref{tab:CompArchis}) show that while  \hyperref[sec:LCAttn]{LCAttn} use far fewer parameters, it fail to match \hyperref[sec:ToM]{ToMMer}’s overall performance. LTQK maintains high recall but loses precision, while LCAttn’s precision drops dramatically. This highlights a clear tradeoff: extreme parameter reduction can preserve recall, but strong precision requires more capacity.

\begin{table}[h]
\centering
\resizebox{\columnwidth}{!}{%
\begin{tabular}{@{}rccccccc@{}}
\toprule
type                          & layer & Agg R & Agg P   & Agg F1 & LLM P & \#params  \\ \midrule
\hyperref[sec:ToM]{ToM}       &  6  & 92.6 & \textbf{23.2}    & \textbf{37.1}   & \textbf{92.5}  & 264 198    \\
\hyperref[sec:LTQK]{LTQK}     &  0   & \textbf{94.4} & 17.6    & 29.7   & 78.8  & 165 894    \\
\hyperref[sec:LCAttn]{LCAttn} &  0-10   & \textbf{94.3} & 9.8    & 17.8    & 44.2  & 2 279      \\ \bottomrule
\end{tabular}%
}
\caption{Comparison of all tested architectures (all models are trained on \textsc{Llama-3.2-1B}). We report the Precision (P) Recall (R) and F1 (F1) aggregated on all 13 tested benchmarks. LLM P is the mean precision on LLM annotated data splits.}
\label{tab:CompArchis}
\end{table}

\section{Supplementary Ablation Studies}\label{app:Ablations}

\subsection{Impact of Rank}
\begin{figure}[t]
\centering
    \includegraphics[scale=.6]{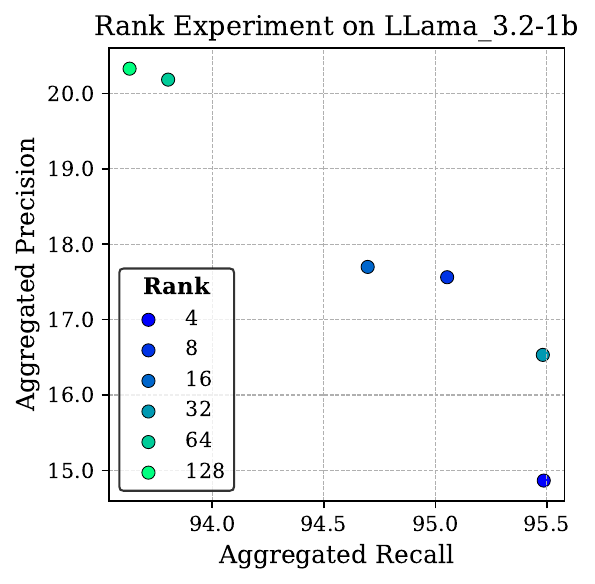}
    \caption{Aggregated precision vs recall of ToMMeR Models accross ranks $r$ as defined in \Cref{sec:ToM}. We choose $64$ for other experiments, yielding a good balance between recall and precision, while maintaining a small number of parameters in ToMMeR.}
    \label{fig:rank_ablation}
\end{figure}

We show \Cref{fig:rank_ablation} the precision-recall balance when varying rank $r$ as defined in \Cref{sec:ToM}, we chose a final rank of $64$ in our other experiments, as it is a good balance between recall and precision, while maintaining a small number of parameters in ToMMeR.

\section{Multi-Head Self Attention Model}\label{sec:OtherArchis}

In  transformer-based language modeling, Multi Head Self Attention (MHSA) is the \cite{vaswaniAttentionAllYou2017} main mechanism used to transfer information between token representations. 
We also tried to use a key and query model using an MHSA layer rather than raw projections, enabling to model contexual cues into queries and keys. We compute the entity span logit probability $m_{ij}$ as : 

\begin{equation}
    m_{ij} = \text{MHSA}_Q(\bz_i) \cdot \text{MHSA}_K(\bz_j)
\end{equation}

However, training was much longer and performance metrics were lower than using the probability in Section~\ref{sec:ToM} of this paper.

\section{Dataset descriptions}
\label{sec:datasets-appendix}
\subsection{MultiNERD}
\label{sec:multinerd}
\textbf{Languages and domains.}  MultiNERD \citep{tedeschi-2022-multinerd}
is a language--agnostic dataset that automatically annotates texts from
Wikipedia and WikiNews in ten languages (Chinese, Dutch, English,
French, German, Italian, Polish, Portuguese, Russian and Spanish).
\textbf{Entity types.}  It provides fine–grained annotation for
15 entity categories (\textit{person}, \textit{location},
\textit{organization}, \textit{animal}, \textit{biological entity},
\textit{celestial body}, \textit{disease}, \textit{event},
\textit{food}, \textit{instrument}, \textit{media}, \textit{plant},
\textit{mythical entity}, \textit{time} and \textit{vehicle}) and
adds disambiguation links to the corresponding Wikipedia pages
\citep{tedeschi-2022-multinerd}.  The annotations are created by
combining WikiNEuRal silver‑data creation and NER4EL fine–grained
labeling, resulting in a high‑quality multi‑genre resource.

\subsection{CoNLL‑2003}
\label{sec:conll2003}
The CoNLL‑2003 shared task \citep{sang2003introductiontoconll2003sharedtask}
provides a benchmark for language–independent named entity
recognition.  It contains English and German newswire
articles taken from Reuters and Frankfurter Rundschau.  Four entity
classes (\textit{person}, \textit{location}, \textit{organization} and
\textit{miscellaneous}) are annotated.  The English portion
comprises 946 training articles, 216 development articles and 231
test articles, while the German portion contains 518 training, 52
development and 342 test articles \citep{sang2003introductiontoconll2003sharedtask}.

\subsection{CrossNER}
\label{sec:crossner}
CrossNER \citep{liu2020crossnerevaluatingcrossdomainnamed}
is a cross‑domain NER dataset covering five specialist domains:
\emph{politics}, \emph{artificial intelligence}, \emph{music},
\emph{literature} and \emph{science}.  Each domain contains a
small labelled training set (100–200 documents) and roughly 1,000
development and test sentences.  Entity types are tailored to each
domain (e.g., \textit{politician}, \textit{election},
\textit{software}, \textit{research field}), and unlabeled domain‑specific
corpora are provided for domain adaptation.  This resource is used
to evaluate whether models generalise across domains with distinct
entity inventories \citep{liu2020crossnerevaluatingcrossdomainnamed}.

\subsection{NCBI Disease corpus}
\label{sec:ncbi}
The NCBI disease corpus \citep{dogan2014ncbi}
contains 793 PubMed abstracts that are fully annotated at both the
mention and concept levels for disease names.  Manual annotation
produced 6,892 disease mentions mapped to 790 unique disease
concepts.  Approximately 88~\% of concepts link to a MeSH entry and
91~\% of mentions correspond to a single concept
\citep{dogan2014ncbi}.  The corpus is split into training,
development and test sets and serves as a benchmark for biomedical
NER and concept normalisation.

\subsection{FabNER}
\label{sec:fabner}
FabNER \citep{kumar2021fabner} is a manufacturing domain corpus
containing over 350{,}000 words of scientific abstracts from the Web of
Science.  Each word is labelled with one of 12 categories covering
materials (\textit{MATE}), manufacturing processes (\textit{MANP}),
machines/equipment (\textit{MACEQ}), applications (\textit{APPL}),
features (\textit{FEAT}), mechanical properties (\textit{PRO}),
characterisation techniques (\textit{CHAR}), parameters
(\textit{PARA}), enabling technology (\textit{ENAT}), concepts or
principles (\textit{CONPRI}), manufacturing standards (\textit{MANS})
and biomedical concepts (\textit{BIOP}) \citep{kumar2021fabner}.
Annotations follow the BIOES tag scheme.

\subsection{WikiNEuRal}
\label{sec:wikineural}
WikiNEuRal \citep{tedeschi2021wikineural} generates silver‑standard NER
training data by combining neural models and the BabelNet knowledge
base.  It produces automatically labelled corpora for nine
languages (Dutch, English, French, German, Italian, Polish, Portuguese, Russian and Spanish) using Wikipedia articles.  The method improves span‑based F1 scores by up to six points over previous approaches for multilingual silver data creation
\citep{tedeschi2021wikineural}.

\subsection{OntoNotes 5.0}
\label{sec:ontonotes}
OntoNotes 5.0 \citep{weischedel2013ontonotes}
is a large multi‑layer corpus containing annotations for syntax,
predicate–argument structure, word sense, coreference and named
entities across English, Chinese and Arabic.  The NER layer defines
18 categories, including \textit{PERSON}, \textit{NORP}, \textit{FACILITY}, \textit{ORGANIZATION}, \textit{GPE},
\textit{LOCATION}, \textit{PRODUCT}, \textit{EVENT}, \textit{WORK OF
  ART}, \textit{LAW}, \textit{LANGUAGE}, \textit{DATE}, \textit{TIME},
\textit{PERCENT}, \textit{MONEY}, \textit{QUANTITY}, \textit{ORDINAL}
and \textit{CARDINAL} \citep{weischedel2013ontonotes}.  The English portion comprises approximately 300~k words of newswire, 200~k words
each of broadcast news and broadcast conversation, 200~k words of web
text and 100~k words of telephone conversations \citep{weischedel2013ontonotes}.  Similar corpora are provided for Chinese and Arabic, making OntoNotes one of the largest resources for multi‑genre NER.

\subsection{ACE 2005}
\label{sec:ace2005}
The Automatic Content Extraction (ACE) 2005 corpus
\citep{ace2005} contains around 1{,}800 documents in English,
Chinese and Arabic drawn from newswire, broadcast news, broadcast
conversation, weblogs, discussion forums and conversational
telephone speech \citep{ace2005}.  Entities are annotated with
seven types—\textit{person}, \textit{organization}, \textit{geo‑political
entity (GPE)}, \textit{location}, \textit{facility}, \textit{vehicle}
and \textit{weapon}—and each entity may have multiple mentions
(names, nominals or pronouns) in a document
\citep{ace2005structures}.  The corpus also includes annotations for
relations and events, but we only use the entity annotations in our
experiments.

\subsection{GENIA}
\label{sec:genia}
The GENIA corpus version 3.02 \citep{kim2004jnlpba}
consists of 2{,}000 MEDLINE abstracts selected using the MeSH
keywords ‘human’, ‘blood cells’ and ‘transcription factors’ and
annotated with a fine‑grained taxonomy of 36 entity classes
\citep{kim2004jnlpba}.  For the JNLPBA shared task the 36 classes
were mapped to five coarse categories: \textit{protein}, \textit{DNA},
\textit{RNA}, \textit{cell line} and \textit{cell type}.  An
additional 404 abstracts were annotated for testing, giving a total of
2{,}404 abstracts with over 100{,}000 tokens
\citep{kim2004jnlpba}.

\section{Schema comparison}\label{app:schema-comparison}

\begin{figure*}[h]
    \centering
    %\hspace{-.6cm} % sinon les marges ACL passent pas...
    \resizebox{\textwidth}{!}{%
    \includegraphics{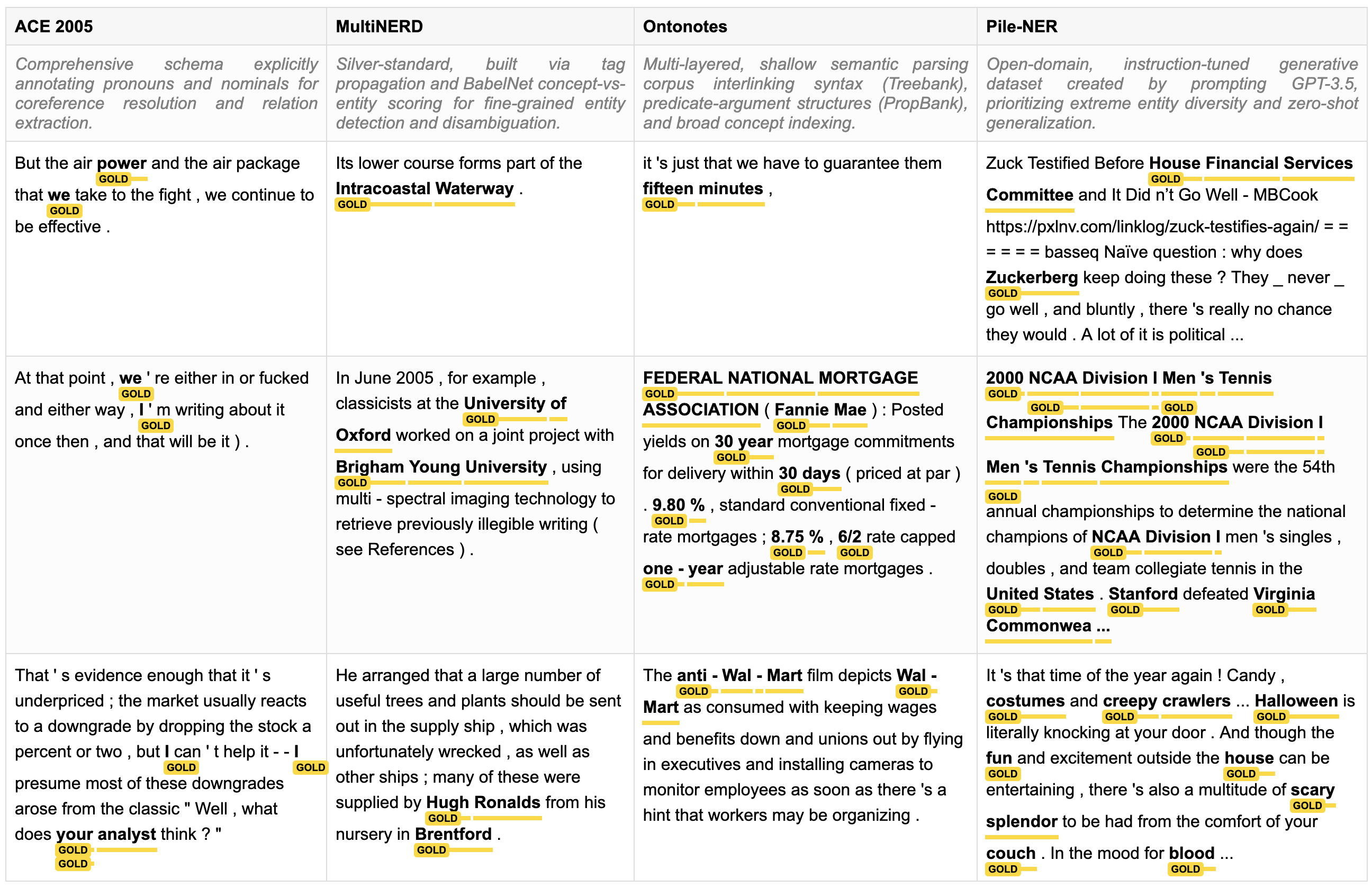}
    }
    \vspace{-0.5cm}
    \caption{A comparative visualization of gold-standard annotations across four distinct benchmarks. The examples illustrate shifting definitions of what constitutes an "entity": ACE 2005 emphasizes structural tracking by tagging pronouns (e.g., "his opponent") and broad nominals (e.g., "nearly everyone") for coreference. MultiNERD focuses on traditional, well-defined proper nouns mapped to fine-grained taxonomy. OntoNotes extends beyond standard names to heavily index nested entities and precise numerical or temporal values (e.g., "fifteen minutes", "9.80 \%"). Finally, Pile-NER demonstrates its generative, open-domain capabilities by extracting highly contextual, non-traditional concepts (e.g., "creepy crawlers", "fun", "blood") alongside complex overlapping spans.}
    \label{fig:SchemaComparison}
    \vspace{-0.5cm}
\end{figure*}

ACE2005, MultiNERD, and PileNER differ mainly in what they treat as an entity mention. ACE2005 uses an entity-centric annotation scheme with a small closed ontology, but a broad notion of mention: it annotates seven core entity types, distinguishes named, nominal, and pronominal mentions, marks the full noun-phrase extent together with a head, and includes mention classes such as specific, generic, negative, and underspecified. As a result, ACE2005 contains many common-noun and pronoun mentions that would not be annotated in standard flat NER. MultiNERD is much closer to conventional named-entity tagging: it uses flat BIO spans and a fixed inventory of 15 fine-grained classes, and its annotations are derived from Wikipedia/Wikinews links plus exact-match/synonym propagation within the document. Compared with ACE2005, it is narrower in mention inclusion because it mainly targets linked surface mentions rather than nominal/pronominal references, but broader in semantic coverage because it adds categories such as ANIM, BIO, CEL, DIS, FOOD, MEDIA, MYTH, TIME, and VEHI. It also explicitly relaxes strict “entity” status for some classes, keeping concept-like items such as animals, plants, foods, and diseases when needed. PileNER differs most sharply from both: it does not use a fixed ontology at all, but an open-world setup in which GPT-3.5 was prompted to “\textit{extract all entities and identify their entity types,}” producing 45,889 examples with about 240k spans and 13,020 distinct type names. We provide an overview of schemas as well a some sampled examples in \Cref{fig:SchemaComparison}.

\end{document}